\documentclass[smallcondensed]{svjour3}

\usepackage[margin={2.5cm,2.5cm}]{geometry}
\usepackage{amsmath}
\usepackage{amsfonts}
\usepackage{amssymb}
\usepackage{mathtools}
\usepackage{bm}
\usepackage{url}
\usepackage{multirow}
\usepackage{latexsym}
\usepackage{booktabs}
\usepackage{algorithm}
\usepackage{algpseudocode}
\usepackage{refcount}

\usepackage{natbib}
\DeclareMathOperator*{\argmin}{argmin}

\title{Interpretable Time Series Classification using Linear Models and Multi-resolution Multi-domain Symbolic Representations}
\titlerunning{Interpretable Time Series Classification using Linear Models}        

\author{Thach Le Nguyen \and Severin Gsponer \and Iulia Ilie \and Martin O'Reilly \and Georgiana Ifrim}

\institute{Insight Centre for Data Analytics, University College Dublin, Ireland \\
           \email{\{thach.lenguyen,severin.gsponer,iulia.ilie,martin.oreilly,georgiana.ifrim\}@insight-centre.org}             \\
            Contact authors:  Thach Le Nguyen, Georgiana Ifrim. 
}
\date{Received: date / Accepted: date}

\begin{document}
\maketitle

\begin{abstract}
The time series classification literature has expanded rapidly over the last decade, with many new classification approaches published each year. 
Prior research has mostly focused on improving the accuracy and efficiency of classifiers, with interpretability being somewhat neglected.
This aspect of classifiers has become critical for many application domains and the introduction of the EU GDPR legislation in 2018 is likely to 
further emphasize the importance of interpretable learning algorithms. Currently, state-of-the-art classification accuracy is achieved with very complex models 
based on large ensembles (COTE) or deep neural networks (FCN). These approaches are not efficient with regard to either time or space,  
are difficult to interpret and cannot be applied to variable-length time series, requiring pre-processing of the original series to a set fixed-length.
In this paper we propose new time series classification algorithms to address these gaps.
Our approach is based on symbolic representations of time series, efficient sequence mining algorithms and 
linear classification models. Our linear models are as accurate as deep learning models but are more efficient regarding running time and memory, can work with variable-length time series 
and can be interpreted by highlighting the discriminative symbolic features on the original time series.
We advance the state-of-the-art in time series classification by proposing new algorithms built using the following three key ideas: 
(1) \textbf{Multiple resolutions of symbolic representations:} 
we combine symbolic representations obtained using different parameters, rather than one fixed representation (e.g., multiple SAX representations);  
(2) \textbf{Multiple domain representations:} 
we combine symbolic representations in time (e.g., SAX) and frequency (e.g., SFA) domains, to be more robust across problem types;  
(3) \textbf{Efficient navigation in a huge symbolic-words space:}
we extend a symbolic sequence classifier (SEQL) to work with multiple symbolic representations 
and use its greedy feature selection strategy to effectively filter the best features for each representation.
We show that our multi-resolution multi-domain linear classifier (mtSS-SEQL+LR) achieves a similar accuracy to the state-of-the-art COTE ensemble, and 
to recent deep learning methods (FCN, ResNet), but uses a fraction of the time and memory required by either COTE or deep models. 
To further analyse the interpretability of our classifier, we present a case study on a human motion dataset collected by the authors.
We discuss the accuracy, efficiency and interpretability of our proposed algorithms and release all the results, source code and data to encourage reproducibility.
\end{abstract}

\section{Introduction}
\label{sec:intro}


State-of-the-art time series classification (TSC) accuracy is currently achieved with complex models such as large ensembles (COTE) \citep{bagnall2015time,bagnall2016great,hive-cote} and deep neural networks (FCN, ResNet) \citep{zwang.dnn,IsmailFawaz2018deep}. 
Although these algorithms achieve high accuracy, they are not efficient in either time or space \citep{bagnall2016great}, do not work on variable-length time series and are not interpretable.
Simpler distance-based methods such as 1-Nearest Neighbour with Dynamic Time Warping (DTW) distance \citep{wang:dmkd13} provide strong baselines, but they lag significantly behind in accuracy,  
suffer from high running times and are negatively affected by noise \citep{schaefer:dmkd16}. It is also not possible to interpret the classification decision: 
we can compare the unlabelled time series to a labelled nearest neighbour, but we get no information about which parts of the time series are important for the classification decision. 

Recent dictionary-based methods built on top of symbolic representations of time series deliver promising accuracy. 
The Symbolic Aggregate Approximation (SAX) by \cite{lin-sax:dmkd07} 
has been the inspiration of numerous studies on time series analysis such as \cite{castro:motif,chen:palmprint,kasten:acoustic}, and \cite{costa:patterns}. 
SAX is a discretisation of time series in the time domain, and produces a sequence of symbols representing the raw numeric data.
It allows compression and enables data mining algorithms to run efficiently. SAX-VSM \citep{senin-saxvsm:icdm13}, FastShapelet \citep{fast-shapelet:sdm13} and SAX-SEQL \citep{fvseql:7930038} 
are representative time series classification methods based on SAX. Existing SAX-based methods are highly dependent on SAX parameters and do not achieve state-of-the-art accuracy \citep{fvseql:7930038}.
A new symbolic representation introduced by~\cite{Schafer:2012:SSF:2247596.2247656}, named Symbolic Fourier Approximation (SFA), and its classification frameworks (BOSS, WEASEL and MUSE) \citep{schafer2015boss,schaefer:dmkd16,Schafer:2017:weasel,muse}, have further advanced the state-of-the-art regarding both accuracy and efficiency. 
SFA uses discretisation in the frequency domain, and by itself does not provide significantly enhanced accuracy as compared to SAX\footnote{\label{Firstfootnote}{According to our experiments comparing single SAX vs single SFA classifiers, shown in Section \ref{sec:method}.}}. 
To achieve high accuracy, the proposed SFA-based algorithms BOSS and BOSS VS \citep{schafer2015boss,schaefer:dmkd16} 
ensemble many models built on different SFA representations. The WEASEL  approach \citep{Schafer:2017:weasel} creates a very large feature space 
using multi-resolution SFA, applies feature selection and then learns a linear model. This method has high accuracy, but it is not memory-efficient. 
Furthermore, it only uses SFA features, which is not appropriate for some problem types\footnote{Experiments and discussion backing these statements are available in Section \ref{sec:experiments} and Section \ref{sec:casestudy2}.}. 

While symbolic representations hold promise, the process of finding the optimal symbolic representation is costly. The SAX-VSM algorithm \citep{senin-saxvsm:icdm13} attempts to find the optimal SAX transformation parameters with an optimization algorithm (DIRECT) and an evaluation method (cross-validation). 
However, it fails to outperform the accuracy of recent state-of-the-art classifiers, e.g., BOSS, WEASEL, COTE.
State-of-the-art classifiers such as BOSS \citep{schafer2015boss} and WEASEL \citep{Schafer:2017:weasel} 
generate symbolic representations at multiple resolutions by varying the symbolic representation parameters (e.g., working with 
multiple SFA representations obtained with different fixed parameters).
This shotgun approach, despite its simplicity, showed to be more robust and efficient than globally optimising the parameters of one fixed representation.
Another motivation for multiple representations of time series is the possibility of exploring different representation domains (e.g., time and frequency domain). 
Representations from different domains describe time series from different perspectives and by combining this knowledge it is possible to train a better model. 
An ensemble like COTE \citep{bagnall2015time} takes advantage of knowledge on time series extracted from different domains. 
However, COTE gathers knowledge provided by its member classifiers, which have a combined time complexity that makes this approach impractical for 
many real-world problems. Symbolic representations are appealing as they can standardize the representations in a sequential structure, 
effectively unifying knowledge from different domains without the need of multiple classification algorithms.
Existing approaches mostly combine the same symbolic representation obtained at multiple resolutions, 
but do not take advantage of distinct symbolic representations from multiple domains, possibly because this would generate a very large feature space.
The key challenge is to develop a method that is able to efficiently work with a huge feature space of symbolic words obtained from distinct symbolic representations, 
and to select the best feature subset without having to explicitly evaluate each feature.

We build on these findings to propose two new time series classification algorithms based on multi-resolution multi-domain symbolic representations.
The first algorithm achieves high accuracy by ensembling models built on multiple symbolic representations. The 
 second algorithm achieves even better accuracy by (i) creating a large feature space using multiple symbolic representations, 
 (ii) efficiently selecting important symbolic features and (iii) training a linear model (logistic regression). 
 For efficient training using a massive symbolic feature space, we extend a symbolic sequence learning algorithm named SEQL~\citep{ifrim-seql:kdd11}.
 SEQL was originally designed as a binary classifier for sequence data such as DNA or text. It works on either
 sequences of symbols (no white space) or sequences of words. The algorithm is  able to explore the all-subsequence space 
 by employing a branch-and-bound feature search strategy and it can efficiently train a linear model based on selected discriminative subsequences.
 In~\cite{fvseql:7930038} we proposed a time series classification algorithm that combines one fixed SAX representation with the SEQL learning algorithm. 
 That approach works with variable-length SAX words which makes the SAX representation less dependent on parameters, but 
 its accuracy is only comparable to SAX-VSM and lags behind the accuracy of more recent classifiers, e.g., COTE and FCN. 
Aiming to improve the accuracy of the combo of symbolic representations and simple linear models, we extend SEQL to work with both SAX and SFA symbolic representations, 
 and to extract features from multiple resolutions and multiple domains. 
 Our approach can also be extended to include other symbolic representations that turn the original time series into a sequence of symbols.
 We explain the requirements for the  candidate symbolic representations in Section \ref{sec:method}.

In our experiments, we explore different combinations for the input representation: a single symbolic representation with fixed parameters (e.g., SAX with fixed parameters), 
a single symbolic representation with multiple parameters (e.g., SAX with multiple resolutions) and distinct symbolic representations with multiple parameters  
(e.g., combining SAX and SFA representations and varying their parameters). 
We evaluate our algorithms on the well-known UCR time series classification benchmark \citep{ucr:keogh15} and discuss the impact  
of using different symbolic representations and learning algorithms on different types of problems (e.g., motion, image, sensor, ECG time series). 

We also investigate another aspect of TSC which is the interpretability of the model. The ability to explain a classification decision is often excluded from the discussion in this field, 
as the community has focused mostly on classifier accuracy and efficiency. 
However, while accuracy and efficiency are very important, the model interpretability is an essential evaluation concern for any time series classifier. 
In many applications, it is important to know the key parts of the input data which are relevant for the analysis task or to understand the classification decision. 
For example, in sports science, an athlete executes a particular exercise and expects automated feedback on whether the exercise is executed correctly or not, and when it 
is incorrect, the athlete should receive feedback on which parts of the movement need correction.
For the TSC problem, we want to highlight to the users the data examined by the model in order to make predictions. 
Our main interpretable classifier is a linear model (i.e., a list of weighted features), so we can use the weighted features learned 
by the model to highlight the parts of the time series that lead to a classification decision. 
We present a case study on TSC interpretability using a human motion dataset and a discussion by a domain expert in jump technique biomechanics (Section \ref{sec:casestudy2}).

\textbf{Our main contributions are as follows:}
\begin{itemize}
\item We present new TSC algorithms which incorporate symbolic representations from multiple resolutions and multiple domains (SAX and SFA) by extending an efficient sequence learning algorithm (SEQL).
\item We analyze the theoretical time and space complexity of all our proposed algorithms.
\item We present an extensive experimental study of our approaches on the UCR Time Series Archive and compare to the state-of-the-art TSC methods.
\item We demonstrate how to interpret our linear classification models in the context of time series analysis on known datasets. 
\item We conduct a case study on the interpretability of our TSC models for a human motion dataset. We also record, report and discuss 
the accuracy, running time and memory usage of all the classifiers evaluated on this real-world problem.
\item All our code, data and detailed results are available from \url{https://github.com/lnthach/Mr-SEQL}.
\end{itemize}

The rest of this paper is structured as follows. In Section \ref{sec:relwork} we describe related prior work. In Section \ref{sec:prelims} we describe the requirements for the input symbolic representations and describe SAX and SFA in detail.
In Section \ref{sec:method} we present our proposed TSC framework. In Section \ref{sec:experiments} we present experimental results. In Section \ref{sec:interpretability} we discuss 
model interpretation in the context of TSC. In Section \ref{sec:casestudy2} we present a case study for the interpretability of our classifier on human motion data. In Section \ref{sec:conclusion} we conclude and discuss future work directions.

\section{Related Work}
\label{sec:relwork}


Nearest neighbour classifiers with Euclidean or time warping distances are typical baselines for time series analysis. \cite{wang:dmkd13} studied 8 different time series representations and 9 similarity measures and evaluated their performance on 38 datasets across various domains and tasks. 
In particular, for the TSC task, the study used 1-Nearest-Neighbor (1NN) classifiers to evaluate the accuracy of these measures. 
The conclusions provide interesting insights into the effectiveness of these measures and reaffirm the competitiveness of DTW in comparison to newer methods.
A more recent comparison of elastic distance measures is presented in a study by \cite{Lines2015:ee} where an ensemble approach was shown to outperform a strong DTW baseline (where the warping window, a parameter important for DTW, was set through cross-validation).
Nevertheless, elastic distance based approaches, even those based on ensembles, are outperformed in accuracy by more recent approaches.

Recently, there has been notable interest in shapelet-based classification algorithms after the first proposal by \cite{ye-shapelets:kdd09}. 
Shapelets are discriminative segments extracted from time series and can be used for classification. 
Moreover shapelets are interpretable, thus they can offer insight into the data. 
Since shapelet discovery is usually time-consuming, studies by \cite{Ye2011,fast-shapelet:sdm13,Gordon_shapelet} 
focused on enhancing the efficiency of the process. FastShapelet \citep{fast-shapelet:sdm13} finds $k$-best shapelets in the dimension-reduced space of SAX. 
It speeds up shapelet discovery, but its accuracy is comparable to \cite{ye-shapelets:kdd09}. 
On the other hand, \cite{grabocka-lts:kdd14} formulated the problem 
as an optimization task and solved it with a stochastic gradient learning algorithm. 
The studies \citep{Lines:2012:STT:2339530.2339579,biST2015} used the shapelets to create a transformed dataset, 
in which the distance between the time series and a shapelet is a feature. One challenge with shapelet-based algorithms, e.g., FastShapelet, is that repeated runs of the 
algorithm produces different shapelets, which also affects the interpretation.

COTE, first introduced as Flat-COTE~\citep{bagnall2015time}, is an ensemble method which incorporates 35 different classifiers for TSC.
HIVE-COTE~\citep{hive-cote}, a recent extension of COTE, added 3 more classifiers to the collection. 
Overall, COTE is among the most accurate TSC algorithms that have been tested on the UCR benchmark \citep{ucr:keogh15}. 
It is one of the few classifiers that incorporate descriptions of time series from different domains. However, its learning framework is based on a large ensemble. This demands substantial computation resources as COTE's time complexity is determined by the slowest algorithm. The work by \cite{Kate2016} is another example of a multi-domain classifier: the algorithm combines SAX and DTW feature spaces for classification.

The excellent recent survey on TSC \citep{bagnall2016great} has contributed a systematic framework to evaluate time series classifiers.
In this study, the authors reproduced the experiments of 18 state-of-the-art classifiers in addition to two baseline classifiers (1-NN DTW and Rotation Forest)
 on the extended UCR benchmark that includes 85 datasets across a range of different TSC problem types (e.g., motion, image, ECG). 
The results were analysed based on algorithm type and problem type. The main claim of the survey was that benchmark classifiers are hard to beat and that COTE was by far the most accurate algorithm. 
Nevertheless, the authors mainly focused on the classifier accuracy for evaluation, and did not evaluate either the efficiency or interpretability of the methods compared.


Regarding symbolic representations of time series, SAX is perhaps the most studied representation \citep{lin-sax:dmkd03,lin-sax:dmkd07,Lin2012:iis,fast-shapelet:sdm13}. The BOP (Bag-Of-Patterns) approach presented by \cite{Lin2012:iis} builds a histogram of SAX words for each time series. A new sample is classified by comparing the histograms to find the nearest neighbour in the training set. SAX-VSM~\citep{senin-saxvsm:icdm13} is another SAX-based classifier which also uses the BOP framework. It first builds a dictionary of distinct SAX words from training data (a vector space representation) and for efficiency reasons, instead of histograms it computes a single vector of tf-idf weights for each class. In addition, it employs an optimization algorithm to search for the optimal parameters of SAX. However the tuning cost is substantial due to the need of cross-validation. In addition, the SAX-VSM authors have analysed the interpretability of SAX-VSM models by mapping SAX words having high tf-idf scores back to the original time series. Our prior work~\citep{fvseql:7930038} proposed a novel TSC algorithm using symbolic representations.
Our approach was a combination of a fixed symbolic representation (SAX) and two adaptations of a sequence classifier (SEQL by~\cite{ifrim-seql:kdd11}). 
One of the proposed classifiers, SAX-VFSEQL, learns approximate subsequences of symbolic words and can thus reduce the influence of SAX parameters and noise on the classification accuracy.
Although more flexible than previous approaches that cannot learn symbolic sub-words, 
 the accuracy of SAX-VFSEQL still suffers from being limited to only one fixed SAX representation and thus falls behind the accuracy of more recent methods.

\cite{Schafer:2012:SSF:2247596.2247656} introduced a new symbolic representation to index time series, the Symbolic Fourier Approximation (SFA). This approach 
uses a Discrete Fourier Transform as the core approximation technique. Based on this work, the authors proposed several classification frameworks for time series, 
which includes 1NN-BOSS  \citep{schafer2015boss} and BOSS VS  \citep{schaefer:dmkd16} (ensemble methods), WEASEL  \citep{Schafer:2017:weasel}, and MUSE \citep{muse}. 
BOSS uses ensembles of histograms of SFA-words and 1NN classifiers, while BOSS VS uses tf-idf class centroids and 1NN for classification.
In the SFA-based TSC algorithms family, WEASEL is the most recent work on univariate time series, while MUSE was developed to classify multivariate data. 
Both methods employ heavy feature engineering and feature selection techniques to filter the huge feature space created by multiple resolutions of SFA transformations, before feeding the 
selected features to a linear model. WEASEL is more accurate than the BOSS algorithms, 
but suffers from memory efficiency issues since it does not include effective methods for pruning features early, and hence also needs to 
carefully restrict the feature space (e.g., by restricting the SFA parameters and the type of features).
The authors of WEASEL and MUSE do not discuss the interpretability of these methods, arguably because of the non-linear characteristics of the SFA transformation.
As we show in Section \ref{sec:casestudy2}, our algorithms are more accurate than WEASEL and use one order of magnitude less memory for training models.
 
The popularity surge of deep learning has inspired various studies to exploit its power for TSC. A comprehensive review of state-of-the-art deep learning algorithms for TSC was published recently~\citep{IsmailFawaz2018deep}. The studied algorithms include: Multi Layer Perceptron (MLP), Fully Convolutional Neural Network (FCN), Residual Network (ResNet), Encoder, Multi-scale Convolutional Network (MCNN), Time Le-Net (t-LeNet) and a few others.
Similarly, the study by~\cite{zwang.dnn} examined the performance of FCN, ResNet, and MLP on the UCR Archive. 
We compare our proposed TSC algorithms to these deep learning approaches and discuss their accuracy as reported by \cite{IsmailFawaz2018deep}. 
Although \cite{zwang.dnn} emphasize the simplicity of their method and its effectiveness as a TSC baseline, there is no detailed discussion about efficiency.
We downloaded the code but we could not reproduce the experiments on a regular PC in a reasonable amount of time, as we did with many of the other existing algorithms. On the other hand, \cite{IsmailFawaz2018deep} outlined the huge amount of effort and computational resources (8,730 experiments, 60GPUs, 100 days of running time) required for their study of deep learning approaches.
Additionally, these approaches do not work on variable-length time series and thus require pre-processing of the original dataset to create fixed length time series. In comparison, our algorithms are as accurate as large ensembles (Flat-COTE, HIVE-COTE) and deep learning approaches (FCN, ResNet), only need a few hours to train and predict for the entire UCR archive on a regular PC, and use orders of magnitude less memory. We provide a detailed discussion of the accuracy and time/memory efficiency of the TSC methods compared in Section \ref{sec:experiments} and Section \ref{sec:casestudy2}.

\section{Symbolic Representation of Time Series}
\label{sec:prelims}

In this section, we discuss two notable symbolic representations of time series: 
the Symbolic Aggregate approXimation (SAX) \citep{lin-sax:dmkd03} and the Symbolic Fourier Approximation (SFA) \citep{Schafer:2012:SSF:2247596.2247656}. 
Although they are different techniques that generate descriptions of time series in different domains, SAX and SFA produce very similar output in terms of structure. 
This property is strongly desirable in our approach as our classifiers require that the symbolic representation, regardless of the method, has a certain standard structure. 
Moreover, both have been shown to be powerful representations for the TSC task.

Generally, the output of both techniques can be described as a sequence of symbols taken from an alphabet $\alpha$, e.g., \textit{aaba}. 
In practice, it is common to employ a sliding window of length $l$ and repeatedly apply the transformation on the time series within this window. 
As a result, the output is a sequence of words, each of which is actually a symbolic sequence of length $w$, e.g., for $w=4$, \textit{abba abbc bacc aacc}.

Formally, a symbolic sequence $S$ of length $n$ has the following form:

\begin{center}
$S$ = $s_1 s_2 \dots s_n$ where $s_i$ is in $\lbrace a_1,a_2, \dots a_{\alpha}\rbrace \cup \lbrace \textvisiblespace \rbrace$
\end{center}
The space character means that S can be either a sequence of symbols or a sequence of words. We will show that the output of SAX and SFA can take the above form and therefore can work with 
our sequence classifier. 
 Table~\ref{table:notation-tsc} summarises the notations used in this paper for the symbolic representations. 
\begin{table}[tbh]
\begin{center}
 \caption{Notation for our Time Series Classification (TSC) framework.}
  \label{table:notation-tsc}
\renewcommand{\arraystretch}{1.5}
\begin{tabular}{ll}\hline
Symbols & Description\\
\hline
$V$ & Raw (normalized) numeric time series\\
$N$ & Number of time series\\
$L$ & Length of original time series\\
$w$ & Length of a symbolic word \\
$\alpha$ & Size of alphabet \\
$l$ & Size of sliding window \\\hline
\end{tabular}
\renewcommand{\arraystretch}{1}
\end{center}
\end{table}

\subsection{Symbolic Aggregate approXimation}

SAX was introduced by \cite{lin-sax:dmkd03} and is a transformation method to convert a numeric sequence (time series) to a symbolic representation, 
i.e., a sequence of symbols with a predefined length $w$ and an alphabet of size $\alpha$. 
Generally, the technique includes three steps: 
\begin{enumerate}
\item Compute the Piecewise Aggregate Approximation (PAA)~\citep{PAA2001} of the time series.
\item Compute the lookup table for the given alphabet.
\item Map the PAA to a symbolic sequence by using the lookup table.
\end{enumerate}
\begin{figure}[!ht]
\centering
\includegraphics[width=0.8\textwidth]{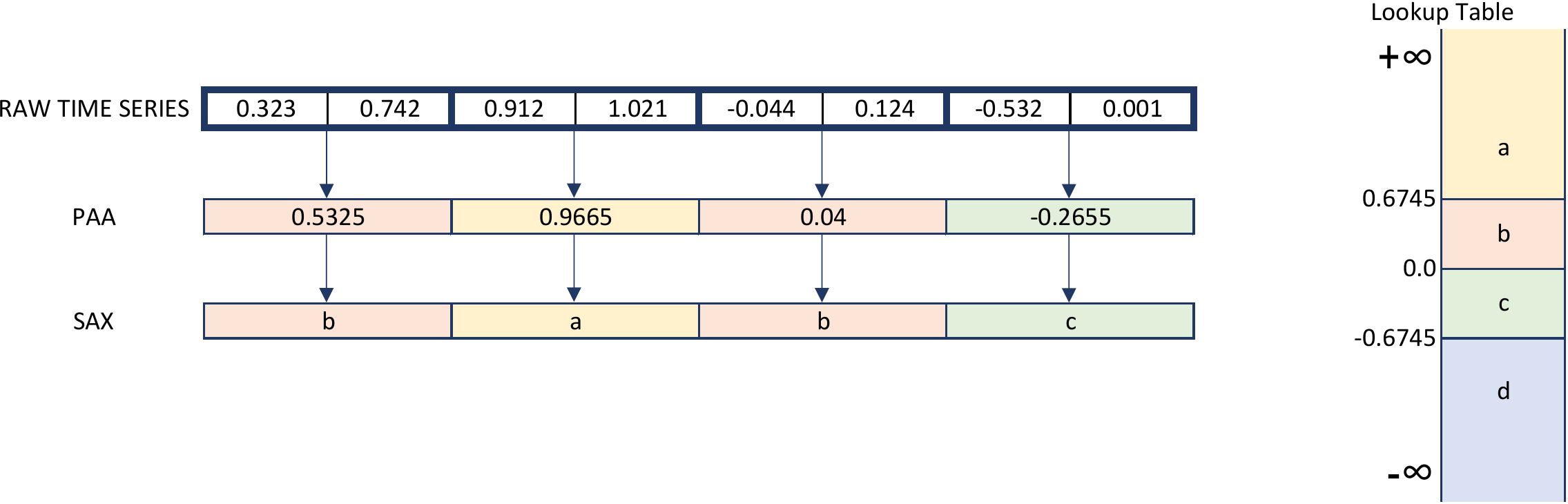}
\caption{An example of SAX transformation that includes: the raw time series, the Piecewise Aggregate Approximation, the lookup table and the final SAX output for $(w=4,\alpha=4)$ .}
\label{fig:saxchars}
\end{figure}

In the first step, the time series is z-normalized and then divided into $w$ equal-length segments and each segment is replaced with its mean value. The result is the PAA vector of length $w$. In the second step, a lookup table is built for the alphabet $\alpha$. Each symbol in the alphabet is associated with an interval, i.e., a continuous range of values. The intervals are obtained by dividing the domain of the time series to $\alpha$ disjoint areas with equal probability, assuming that the values of the time series are normally distributed (hence the z-normalization). Finally, each entry of the PAA vector is then replaced by a symbol taken from the alphabet by using this lookup table. 

Figure~\ref{fig:saxchars} illustrates an example of SAX output with parameters set to $L = 8$, $w = 4$ and $\alpha = 4$. The lookup table divides the domain of the time series into 4 intervals by defining 3 breakpoints ($-0.6745$, $0.0$, and $0.6745$) and links each interval to a symbol from the alphabet $\{a,b,c,d\}$. Each entry in the PAA vector is the average of the corresponding segment in the raw time series. The SAX sequence is produced by looking up the PAA from the table. The first entry ($0.5325$) falls within the range $[0.0,0.6745)$ thus the first symbol taken is $b$.

\begin{figure}[!ht]
\centering
\includegraphics[width=0.8\textwidth]{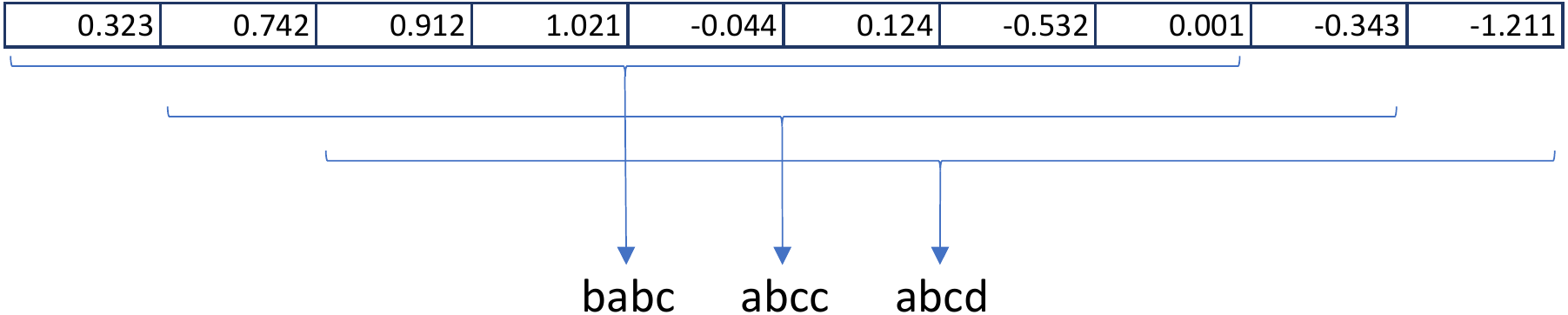}
\caption{Sliding window shifting to obtain a time series representation based on SAX-words $(l=8,w=4,\alpha=4)$ .} 
\label{fig:saxwords}
\end{figure}

SAX can also be combined with a sliding window of length $l$, usually done to process longer time series (Figure~\ref{fig:saxwords}). 
Our previous study \citep{fvseql:7930038} also found that the sliding window technique has a positive impact on the classification accuracy, 
arguably because it can capture a finer description of the time series.

The procedure to transform a time series to a SAX representation with a sliding window can be summarised in Algorithm~\ref{alg:sax_with_sliding_window}. 
The sliding window starts from the first time-stamp, i.e., the beginning of the time series. 
The subsequence of length $l$ within the window is then transformed to a symbolic sequence of length $w$ with the previously described steps. 
This sequence is commonly referred to as a SAX word. The process is repeated until the window reaches the end of the time series. 
Hence the final output is a sequence of equal-length SAX words.

In our implementation of SAX, we also discard continuously repeated SAX words as often done by related works (i.e., numerosity reduction). We apply the same practice on SFA, which will be discussed next.

\begin{algorithm}[h]
\caption{SAX with sliding window}
\begin{algorithmic}[1]
\State Set window length $l$
\State Set word size $w$
\State Set alphabet size $\alpha$
\State Compute lookuptable
\State $L = length(timeseries)$
\ForAll{$t$ in $\left[ 0 , L - l \right) $}
\State $normed\_ts = z\_normalize(timeseries[t:t+l])$
\State $PAA = computePAA(normed\_ts)$
\State $S = ``"$
\For{$v$ in PAA}
\State $S \mathrel{+}= lookup(v)$
\EndFor
\If{$S$ is not a repeated word}
\State Add $S$ to the final representation
\EndIf
\EndFor
\end{algorithmic}
\label{alg:sax_with_sliding_window}
\end{algorithm}

\subsection{Symbolic Fourier Approximation}
\label{subsec:sfa}

SFA \citep{Schafer:2012:SSF:2247596.2247656} also transforms a time series to a symbolic representation. Similarly to SAX, SFA employs a sliding window to extract segments of time series before transformation (although \cite{Schafer:2017:weasel} also suggested non-overlapping window for their SFA-based WEASEL classification framework). Hence SFA's parameters also include the window size $l$, the word length $w$ and the alphabet size $\alpha$. 

The core differences between SAX and SFA are the choices of approximation and discretisation techniques. SFA uses a Discreet Fourier Transform (DFT) method to approximate a time series. 
DFT is well known in the signal processing community and can act as a filter to remove noise from data. 
The same authors also introduced a Multiple Coefficient Binning (MCB) method to discretise the approximation. 
The overall procedure consists of two major steps:

\begin{enumerate}
\item MCB discretisation: Compute the lookup table from the DFT approximations of the training data.
\item Map the DFT approximation of the input time series to its SFA representation with the lookup table.
\end{enumerate}

Both steps employ the DFT technique to approximate an input time series with a vector of length $w$. Basically, DFT decomposes a time series into a series of sinusoid waves, 
each of which can be represented by a Fourier coefficient. The coefficient is a complex number, hence it can be defined by 2 real values: one for the imaginary part and one for the real part. 
Therefore, only the first $w/2$ coefficients of the series are used to create a sequence of length $w$.

\begin{figure}[!ht]
\centering
\includegraphics[width=0.8\textwidth]{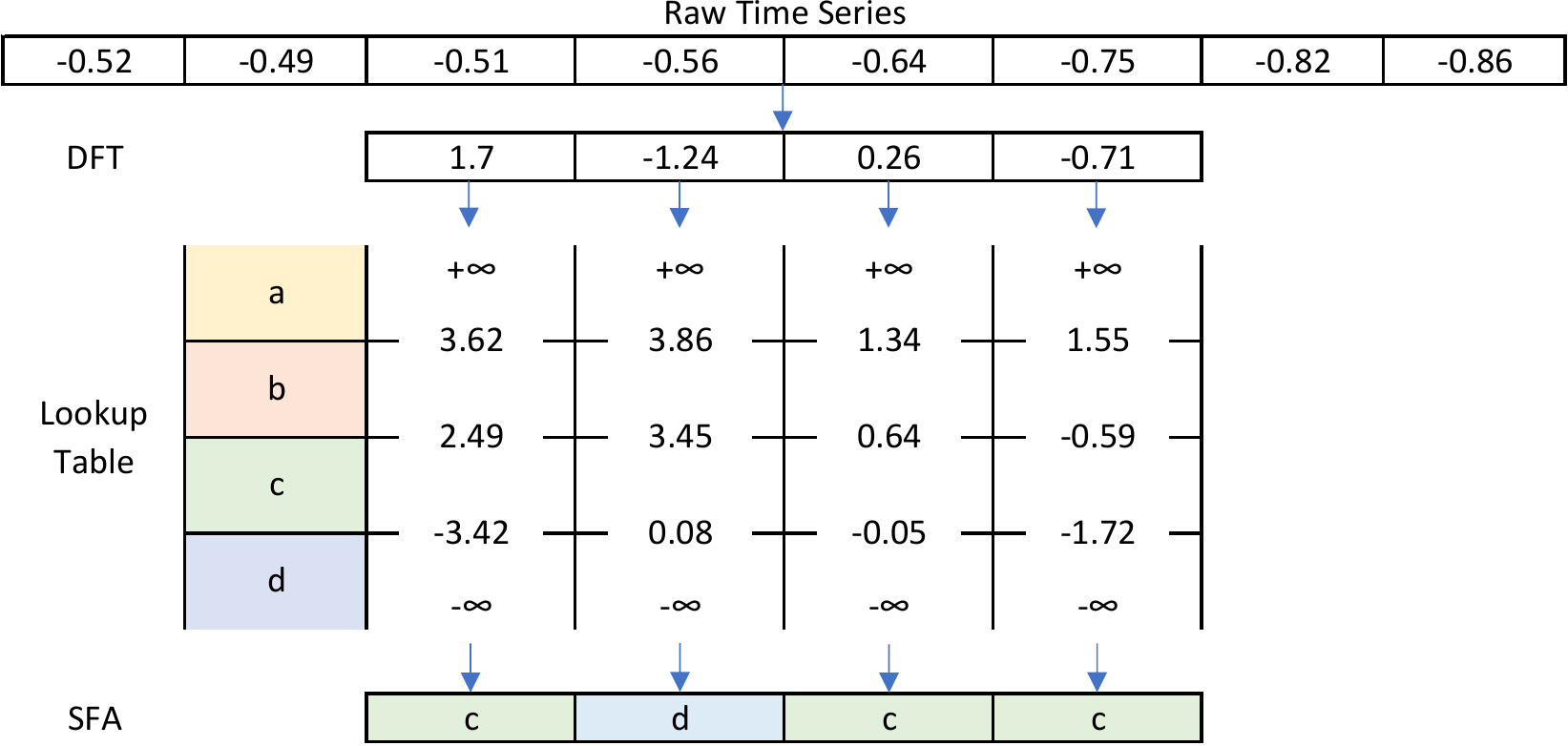}
\caption{An example of SFA transformation that includes: the raw time series, the DFT approximation, the lookup table and the final SFA output for $(w=4,\alpha=4)$ . Note that a symbol has different meanings at each position, e.g., c at position 1 is different from c at position 3, as it represents a different interval for each position.}
\label{fig:sfa-example}
\end{figure}

The discretisation step (MCB) computes the lookup table from the training data. It first computes the DFT approximations of all $l$-length segments extracted from the entire training data to obtain a set of $w$-length vectors. Then from this set of vectors, it computes a set of $\alpha - 1$ breakpoints for each i-th position ($1 \leq i \leq w$) according to the distribution of the i-th values (i.e., equi-depth binning method). The areas divided by the breakpoints are associated with the symbols from the alphabet. The result is a lookup table of $w$ columns and $\alpha$ rows (Figure~\ref{fig:sfa-example}).

The transformation step first computes the DFT approximation for each $l$-length segment of the input time series. For each i-th entry of the approximation vector, the corresponding symbol is looked up from the i-th column of the lookup table. Thus each segment of the time series is transformed to a sequence of length $w$. The final result is a sequence of SFA words extracted from the input time series.

It is important to note that, even at the same level of discretisation $\alpha$ (i.e., number of intervals across the domain), a symbol in SFA has more expressive power than a symbol of SAX. 
While a symbol of SFA can represent $w$ different (usually overlapping) intervals ($w = 4$ as in Figure~\ref{fig:sfa-example}), a symbol of SAX can only represent one single interval.


Figure~\ref{fig:sfa-example} gives an example of SFA transformation. The raw time series is DFT approximated with a vector of length $w=4$. The i-th entry of this vector is then replaced by a symbol according to the i-th column in the lookup table.
Table~\ref{table:sax_vs_sfa} summarizes the difference between the SAX and SFA representations. We use the same SFA implementation as described by \cite{Schafer:2012:SSF:2247596.2247656} and report the time complexity from that paper.

\begin{table}[!ht]
\begin{center}
\caption{Comparison between SAX and SFA symbolic representations of time series.}
\label{table:sax_vs_sfa}
\renewcommand{\arraystretch}{1.5}
\begin{tabular}{llll}
\hline 
Method & Approximation & Discretisation & Complexity \\ 
\hline 
SAX & PAA & equi-prob & $\mathcal{O}(NL\log{}L)$ \\ 
\hline 
SFA & DFT & MCB + equi-depth  & $\mathcal{O}(NL)$ \\ 
\hline 
\end{tabular} 
\renewcommand{\arraystretch}{1}
\end{center}
\end{table}

\section{Sequence Learner with Multiple Symbolic Representations of Time Series}
\label{sec:method}

Typically, the SAX representation is susceptible to how we set parameters, 
i.e., the sliding window size $l$, the word size $w$ or the size of the alphabet $\alpha$. 
Each choice of parameters captures a different structure of the time series which is essential for the classification task. 
One solution for this issue is to search for the optimal parameters, either by a naive grid search or a more complex optimization algorithm
 (e.g., DIRECT as in SAX-VSM \citep{senin-saxvsm:icdm13}).  In our previous work, we mitigate this issue by introducing a new algorithm that can learn discriminative sub-words from a SAX word-based representation~\citep{fvseql:7930038}. However, while it was still competitive at the time of publication, that algorithm has fallen behind most recent state-of-the-art (e.g., WEASEL) in terms of accuracy.

Here we introduce a new approach which uses multiple resolutions and multiple domain representations of time series. 
Fast Shapelets, BOSS and WEASEL are notable classifiers using multiple resolutions, although they only support representations from one domain (via SFA). 
Their competitive results demonstrates the potential of combining multiple symbolic resolutions. 
Our hypothesis is that a single representation of time series, even an optimal one, might be insufficient to capture the necessary structure for the classification task. 
By combining knowledge from multiple resolutions and multiple domains, a learning algorithm can deliver a more robust model.

This section first describes the core technique of our approach, which is a sequence classification framework for the (single) symbolic representation of time series. After that, 
two new methods based on the core technique are proposed, to make use of multiple representations. 
The first one is an ensemble method and the second one is a feature selection method combined with a linear model.

\subsection{Sequence Learner with Symbolic Representation of Time Series}

\begin{figure}[!ht]
\centering
\includegraphics[width=0.7\textwidth]{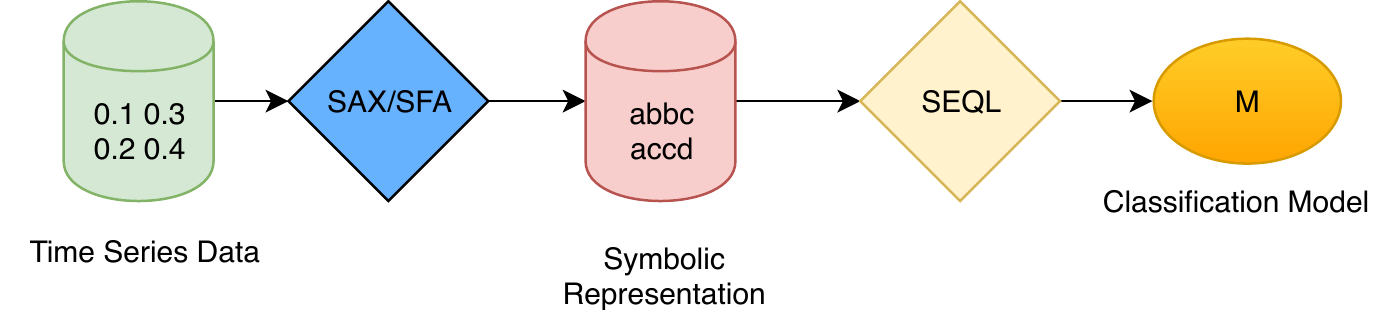}
\caption{SEQL classification algorithm adapted to symbolic representations (SAX or SFA) of time series.} 
\label{fig:sax_vseql}
\end{figure}

Figure~\ref{fig:sax_vseql} sketches our approach which is composed of two components: a symbolic representation and an efficient sequence classifier. 
The symbolic representation can be either SAX or SFA. 

Our core algorithm for classification is Sequence Learner (SEQL) \citep{ifrim-seql:kdd11}. 
SEQL was originally designed as a binary classifier for sequence data such as DNA or text. 
The algorithm is  able to explore the all-subsequence space by employing a branch-and-bound feature search strategy. 
Thus it can select a set of discriminative subsequences in an effective manner. With the symbolic representation of time series, 
the symbolic features can easily be translated back to a set of time series' discriminative segments. TSC with symbolic representation is not a new idea, 
however the most common approach is to build a dictionary directly from the results of the transformation (usually a bag of symbolic words). 

\cite{fvseql:7930038} introduced two adaptations of SEQL for the TSC task. 
The first one (SAX-VSEQL) can learn subsequences from the SAX words while the second one (SAX-VFSEQL) can approximate the subsequences. 
The latter was proposed mainly to make the representation less dependent on the symbolic parameters ($l$, $w$, and $\alpha$). 
As multiple resolutions of a given symbolic representation can create the same effect, we decided to adapt only the lightweight VSEQL version for our new classifiers.
Hence from here on, we use the name SEQL to refer to the VSEQL TSC classifier by \cite{fvseql:7930038}.

The training input for SEQL is a set of sequences and their labels. 
The output is a linear mapping function $f: S\rightarrow  \{-1, +1\}$ to predict the label of new sequences. 
The function $f$ is represented by a parameter vector $\beta = (\beta_1, \dots, \beta_j,\dots, \beta_d)$ (where $d$ is the number of features) which minimizes the loss function $L(\beta)$ using greedy coordinate descent:

\begin{equation}
\label{eq:minL}
\beta^{*}=\argmin_{\beta \in \mathbb{R}^{d}}L(\beta)
\end{equation}
where
\begin{equation}
\label{eq:generalLoss}
L(\beta) = \sum_{i=1}^{N} \xi(y_i,x_i,\beta)+ CR_\alpha (\beta)
\end{equation}
and $\xi(y_i,x_i,\beta)$ is the binomial log-likelihood loss:
\begin{equation}
\label{eq:log-loss}
\xi(y_i,x_i,\beta) = log(1+e^{-y_i\beta^{T} x_i})
\end{equation}

The $x_i$ in Equation (\ref{eq:generalLoss}) and (\ref{eq:log-loss}) denotes the feature vector of the sequence $S_i$ (in all-subsequence feature space) while $y_i$ denotes its true label (either $+1$ or $-1$); $i$ is the index of training examples, $N$ is the total number of training examples. On the right hand side of Equation (\ref{eq:generalLoss}), $C$ denotes the regularization weight and $R_\alpha (\beta)$ denotes the elastic-net regularization. In SEQL the all-subsquence feature space is not explicitly generated; an iterative process is applied to efficiently generate and search the feature space 
for the best feature to be optimized next. The classification decision is computed as $f(x) = \beta^{t} x$. SEQL (binary classifier) was shown to work well for multi-class classification through the one-vs-all approach \citep{ifrim-seql:kdd11}.

\begin{algorithm}[b]
\caption{SEQL workflow}
\label{alg:BasicWorkflow}
\begin{algorithmic}[1]
\State Initialize the subsequence tree with all unigrams
\State Set $\beta = 0$ 
\While{!termination condition}
\State Explore the subsequence tree to find the best subsequence $\hat{s}$ with maximum gradient value //\emph{This step uses branch-and-bound search based on the structure of the feature space; only a small subset of features is generated and evaluated for selecting the best feature, which makes the search very fast. See \cite{ifrim-seql:kdd11} for mathematical details}.
\State Update $\beta$
\EndWhile
\end{algorithmic}
\label{alg:seq}
\end{algorithm}

The SEQL workflow is shown in Algorithm~\ref{alg:seq}. At each iteration, it searches for the coordinate $j$ of $\beta$ which has maximum gradient 
$j = argmax_l \left| \dfrac{\bm{\partial}L}{\bm{\partial}\beta_l}\left( \beta \right) \right|$. Since each coordinate corresponds to a subsequence, this process is translated to finding the  most discriminative subsequence in the context of SEQL.
The search is done by navigating the sequence tree starting from the unigrams. The size of a complete tree is the total number of subsequences, which is normally impractical to traverse. However, SEQL employs a greedy branch-and-bound strategy to selectively generate and prune any unpromising part of the tree. This strategy relies on the anti-monotonicity property of a sequence, i.e., a sequence is always equally or less frequent than all of its subsequences. This property allows the algorithm to calculate a gradient upper-bound for each subsequence, which enables the search-tree pruning decision. 
More details about the algorithm can be found in the original paper~\citep{ifrim-seql:kdd11}.

SEQL takes sequences of characters or sequences of words as input. 
For the TSC task we take as input the sequence of symbolic words resulting from a symbolic transformation of the time series. To be able to 
learn symbolic subwords, we consider the characters (rather than the words) to be the unigrams of the representation, and thus each subsequence of symbols from the training data is a potential feature during learning. 
To achieve this we need to restrict the feature expansion and search to stay within a symbolic word, so the features learned are always contiguous subsequences or subwords.
In our previous paper~\citep{fvseql:7930038}, we discuss more details about this adaptation in the context of the SAX-words symbolic representation.
The output of SEQL training is a linear model which is essentially a set of subsequences and their coefficients (Table~\ref{table:seql_model}). The coefficients can be interpreted as the discriminative power of the subsequence.


\begin{table}[htb]    
	\begin{center}
	\caption{An example model trained by SEQL: a linear model in the space of all symbolic sub-words.}
	\label{table:seql_model}
	\begin{tabular}{lc}
	\hline 
	\\[-0.8em]
	Subsequences & Coefficients \\
	\\[-0.8em] 
	\hline 
	\\[-0.7em]
	cccdbcdda & 0.014 \\ 
	cccdcbcddaaa & 0.013 \\
	ccdbc & 0.012 \\
	cccdbc & 0.007 \\
	ccccdbc & 0.004 \\
	cccbbdd & -0.006 \\
	cccbbd & -0.010 \\
	ccbbddb & -0.011 \\
	cccbb & -0.012 \\
	ccccbb & -0.014 \\
	\\[-0.7em]
	\hline 
	\end{tabular} 
	\end{center}
\end{table}

We discuss next two algorithmic adaptations to be able to combine a single symbolic representation of time series (that produces a sequence of symbolic words from a numeric signal) 
and the sequence mining algorithm SEQL (to efficiently search the entire space of symbolic sub-words and select good features).
The first adaptation, SAX-SEQL, was presented by \cite{fvseql:7930038}. The second adaptation, SFA-SEQL, is proposed here for the first time. 
Based on these adaptations, we then propose two new TSC algorithms: (i) ensemble-based and (ii) linear model-based time series classifiers.

\paragraph{\textbf{SAX-SEQL:}} 
To combine the SAX representation with the SEQL classifier, the time series is transformed to its SAX representation with a sliding window. 
The words are concatenated with a space delimiter to form a sequence of SAX words for each time series. 
Then SEQL learns a linear classification model from the new training data. 
The test data is also transformed to its SAX-words representation before being classified (Algorithm~\ref{alg:sax_seql}).


\begin{algorithm}[h]
\caption{The workflow of SAX-SEQL}
\begin{algorithmic}[1]
\State Set $l,w$ and $\alpha$
\State $train = SAXtransform(train\_time\_series, l,w,\alpha)$
\State $test = SAXtransform(test\_time\_series, l,w,\alpha)$
\State Train with SEQL $M = SEQLearner(train)$. // \emph{This step results in a linear model: a list of weighted symbolic features.}
\State Test with SEQL $predictions = SEQLClassifier(M,test)$
\end{algorithmic}
\label{alg:sax_seql}
\end{algorithm}



\paragraph{\textbf{SFA-SEQL:}}
To combine the SFA representation with the SEQL classifier, the time series is transformed to its SFA-words representation with a sliding window. 
Nevertheless, to be able to use this representation with SEQL, we need to make some modifications to the SFA-words output as described below.
As discussed in Section~\ref{subsec:sfa}, the same symbol in an SFA representation can imply different intervals. 
For example, the lookup table in Figure~\ref{fig:sfa-example} can produce a sequence ``$dabb$" in which the 3-rd and 4-th positions share the same symbol ``$b$". 
However, while the former represents the interval $[0.64,1.34)$, the latter corresponds to interval $[-0.59,1.55)$. 
Moreover, one $b$ symbol is linked to the real part, while the second $b$ is linked to the imaginary part of the same Fourier coefficient. 
Unfortunately, this tricky feature is not compatible with SEQL, since that algorithm would be misled by treating both ``$b$" as the same unigram. 
To fix this issue, we let each column in the lookup table have its own alphabet, e.g., the symbolic representation of the same example would be ``$d_{1}a_{2}b_{3}b_{4}$". 
This way, SEQL can recognize two different unigrams ``$b_{3}$" and ``$b_{4}$". 
Technically, the size of the alphabet for SEQL has increased $w$ times, but the level of SFA discretisation (the number of intervals across the domain) is still $\alpha$. 
This means that SFA does not get more expressive power by this transformation, it is simply a trick on top of standard SFA, to let SEQL know that the semantics of 
the SFA symbols at each position is different. Hence, we still refer to the size of alphabet as $\alpha$ when we discuss the SFA representation.
Besides the above issue, the procedure for training and testing SFA-SEQL is identical to that of SAX-SEQL.

\subsection{Ensemble SEQL}

\begin{figure}[!ht]
\centering
\includegraphics[width=0.8\textwidth]{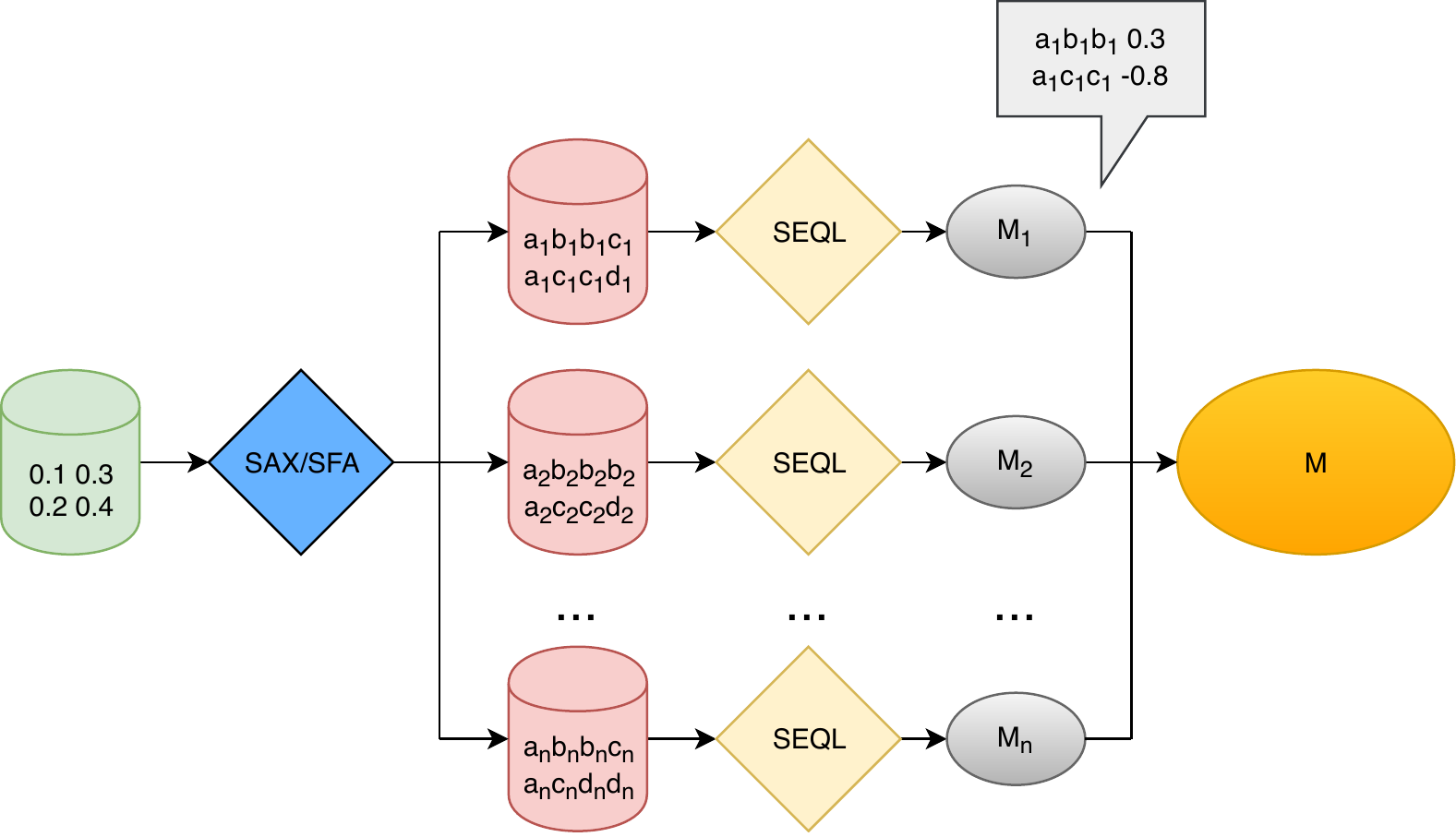}
\caption{An ensemble of SEQL models trained from different SAX (or SFA) representations. 
We call this a multi-resolution approach for a given symbolic representation.} 
\label{fig:ensemble_vseql}
\end{figure}

Ensemble SEQL is a new algorithm we propose to combine SEQL with multiple resolutions and multiple domain symbolic representations. 
Each model is trained by the same classifier (SEQL) but with a different symbolic representation of the training data as input. 
Figure~\ref{fig:ensemble_vseql} illustrates the training procedure to produce $n$ SEQL models from different SAX or SFA representations for the ensemble.
Different representations of the time series can be generated simply by adjusting the transformation parameters ($l$, $w$, and $\alpha$) to obtain multiple resolutions for a given symbolic representation. 
Algorithm~\ref{alg:ensemble_train} shows how we can adjust the window length $l$ to train multiple models (Line 3). As a result, the number of representations is approximately 
$sqrt(L)$, i.e., the longer the time series, the larger the number of representations it can produce. 
For each set of parameters, the raw training data is transformed to the SAX or the SFA representation (Line 4) and a new SEQL model $M_i$ is trained upon this representation (Line 5). 
The output is an ensemble $M$ of all $M_i$.

In our experiments, we fix the values for the word length ($w = 16$) and alphabet size ($\alpha = 4$), and only adjust the sliding window size ($l$) as we observed that there was little benefit in varying all three. 
The minimum window size starts from $minl = 16$, as we should not have a window shorter than the word size, and the step size for increasing the window length is $\sqrt{L}$. It is also worth noting that in the variable length time series scenario, $L$ can be set with the maximum length in the training set. This setting for the window size follows 
similar settings investigated by \cite{schaefer:dmkd16}.
\begin{algorithm}[h]
\caption{Ensemble SEQL: Training}
\begin{algorithmic}[1]
\State Set word size $w=16$
\State Set alphabet size $\alpha=4$
\State Set minimum window size $minl=16$
\For{$l = minl$, $l <= L$, $l \mathrel{+}= sqrt(L)$}
\State $sax = SAXtransform(raw\_time\_series, l,w,\alpha)$
\State Train the SEQL model from the symbolic representation $M_i = SEQL(sax)$
\State $M[l,w,\alpha] = M_i$
\EndFor
\end{algorithmic}
\label{alg:ensemble_train}
\end{algorithm}

\begin{algorithm}[h]
\caption{Ensemble SEQL: Testing}
\begin{algorithmic}[1]
\State Set word size $w=16$
\State Set alphabet size $\alpha=4$
\State Set minimum window size $minl=16$
\State $score = 0$
\For{$l = minl$, $l <= L$, $l \mathrel{+}= sqrt(L)$}
\State $sax = SAXtransform(raw\_time\_series, l,w,\alpha)$
\State $score \mathrel{+}= M[l,w,\alpha].predict(sax)$
\EndFor
\State $prediction = sign(score)$
\end{algorithmic}
\label{alg:ensemble_test}
\end{algorithm}

For the prediction (Algorithm~\ref{alg:ensemble_test}), the unlabelled time series is converted to a SAX representation with the same set of configurations chosen in the training step (Line 6). 
Each model makes a prediction based on the representation of the corresponding configuration (Line 6). The sign of the predicted score aggregation will determine the predicted class of the time series (Line 9).


\subsection{SEQL as Feature Selection}

\begin{figure}[!ht]
\centering
\includegraphics[width=0.9\textwidth]{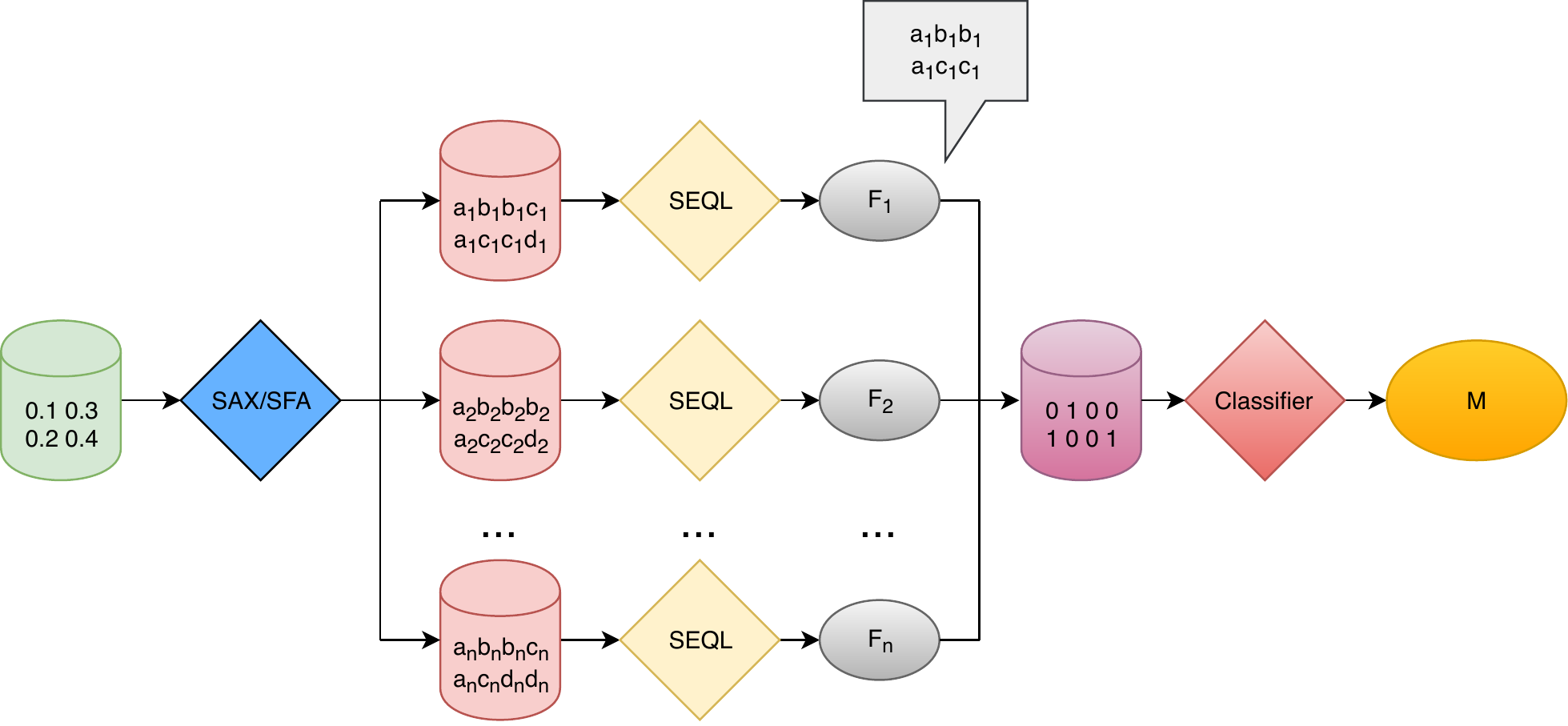}
\caption{SEQL as feature selection method. Features are selected from multiple-resolutions and/or multiple-domain symbolic representations and fed to a logistic regression algorithm.} 
\label{fig:feature_selection_vseql}
\end{figure}

In this section we propose a second new algorithm for training a TSC with multiple symbolic representations and linear models. 
The learning output of SEQL is essentially a list of subsequences selected from the training data (Table~\ref{table:seql_model}), hence the method can be used for feature selection. 
The process diagram for this scheme is illustrated in Figure~\ref{fig:feature_selection_vseql}. 

As it can be seen from the diagram, we  first transform time series data to multiple symbolic representations (either SAX or SFA). 
We then feed each representation to a SEQL trainer. 
Algorithm~\ref{alg:seql_as_fs} explains how the features were extracted using SEQL. 
It is basically Algorithm~\ref{alg:ensemble_train} with an additional for-loop to collect the new features. 
A feature is identified by a subsequence learned by SEQL and the associated SAX (or SFA) configuration (Line 9).
With the collected set of features $F$, it is possible to apply any traditional classification technique such as logistic regression, support vector machine or random forest. 
After experimenting with several learning algorithms, we chose logistic regression for its simplicity, accuracy and interpretability.

\begin{algorithm}[h]
\caption{SEQL as Feature Selection}
\begin{algorithmic}[1]
\State Set word size $w=16$
\State Set alphabet size $\alpha=4$
\State Set minimum window size $minl=16$
\State Set of features $F=\{\}$
\For{$l = minl$, $l <= L$, $l \mathrel{+}= sqrt(L)$}
\State $sax = SAXtransform(raw\_time\_series, l,w,a)$
\State Train the SEQL model $M_i = SEQL(sax)$
\ForAll{$subsequence$ in $M_i$}
\State $F$.add(new Feature($subsequence,l,w,\alpha$))
\EndFor
\EndFor
\end{algorithmic}
\label{alg:seql_as_fs}
\end{algorithm}

It is also worth to note that feature engineering methods are also applicable here. In fact, multiple representations can lead to strongly correlated features since they are essentially generated from the same time series (with different transformations). An extra step to filter unnecessary features might be useful in practice. However, we demonstrate in our experiments that, even with the absence of such filters, the model learnt upon the SEQL feature selection is very accurate on test data.

\subsection{SEQL with Multiple Representations from Multiple Domains}

So far we have discussed the algorithms in the scenario of single-type representations, i.e., either SAX or SFA. However, as SEQL can work with both representations, making use of both for classification is fairly straightforward. The ensemble $M$ can contain both SAX and SFA models: $M = M_{SAX} \cup M_{SFA}$. The set of features $F$ can also contain both SAX and SFA features: $F = F_{SAX} \cup F_{SFA}$.

Table \ref{table:exp_combi} summarizes different combinations between the symbolic representations and all proposed variants of SEQL-based algorithms. We prefix the algorithms that use multiple representations with 'mt'.

\begin{table}[h]
\begin{center}
\caption{Combinations of symbolic representations and variants of SEQL-based algorithms.}
\label{table:exp_combi}
\renewcommand{\arraystretch}{1.5}
\begin{tabular}{llll}
\hline 
Input Type & Symbolic Representation & SEQL variant & Name \\ 
\hline 
Single & SAX & SEQL & SAX-SEQL \\ 
Single & SFA & SEQL & SFA-SEQL \\ 
\hline 
Multiple & SAX & Ensemble SEQL & mtSAX-SEQL \\ 
Multiple & SFA & Ensemble SEQL & mtSFA-SEQL \\ 
Multiple & SAX and SFA & Ensemble SEQL & mtSS-SEQL \\ 
\hline 
Multiple & SAX & SEQL as Feature Selection & mtSAX-SEQL+LR \\ 
Multiple & SFA & SEQL as Feature Selection & mtSFA-SEQL+LR \\ 
Multiple & SAX and SFA & SEQL as Feature Selection & mtSS-SEQL+LR \\ 
\hline 
\end{tabular} 
\renewcommand{\arraystretch}{1}
\end{center}
\end{table}



\subsection{Time and Space Complexity}

Table~\ref{table:complexity} summarize the theoretical complexity of our methods in comparison with well-known time series classifiers. The time complexity of SEQL is proportional to the number of the subsequences it has to evaluate. In the worst case, SEQL has to explore the complete subsequence space, i.e., when it fails to prune any part of the tree. Let  $N_s$ denote the total number of sequences and $l_s$ denote the length of the sequence. The time complexity of SEQL-based classifiers is then:

\begin{align}
T(\text{SEQL})
		& = \mathcal{O}(N_s(1 + \dots + l_s)) \nonumber \\
		& = \mathcal{O}(N_s l_s^2) \label{eq:seql_worst} \
\end{align}

For the symbolic representation of time series, in our algorithm each word is counted as a sequence, accordingly $N_s \leq N(L - l)$ and $l_s = w$:

\begin{align}
T(\text{SAX-SEQL})
		& = T(\text{SAX}) + T(\text{SEQL}) \nonumber \\
		& = \mathcal{O}(NL\log{}L) + \mathcal{O}(N (L-l) w^2) \nonumber \\
		& = \mathcal{O}(NL\log{}L) + \mathcal{O}(N L w^2) \label{eq:sax_seql_1} \
\end{align}

In practice, $w$ is often a constant, hence:

\begin{align}
T(\text{SAX-SEQL})
		& = \mathcal{O}(NL\log{}L) + \mathcal{O}(N L) \nonumber \\
		& = \mathcal{O}(NL\log{}L) \label{eq:sax_seql_2} \
\end{align}

If multiple representations are used, the complexity also depends on the number of representations, which is approximately $\sqrt{L}$:

\begin{equation}
\label{eq:mtsax_seql}
T(\text{mtSAX-SEQL}) = \mathcal{O}(NL^{\frac{3}{2}}\log{}L)
\end{equation}

Similarly, if the type of the representation is SFA:

\begin{equation}
\label{eq:sfa_seql}
T(\text{SFA-SEQL}) = \mathcal{O}(NL)
\end{equation}

\begin{equation}
\label{eq:mtsfa_seql}
T(\text{mtSFA-SEQL}) = \mathcal{O}(NL^{\frac{3}{2}})
\end{equation}

In terms of memory, SEQL needs space to store the training data and the subsequence tree. Generally, the size of the tree is bounded by the alphabet and the length of the subsequences. 
The length of the subsequences in turn is bounded by the word length parameter $w$. 
As a result, the number of nodes in the worst case is $\alpha^w + \alpha^{w - 1} + \dots + \alpha = \dfrac{\alpha^{w+1} - \alpha}{\alpha - 1}$ 
which is nevertheless a constant since $\alpha$ and $w$ are fixed. 
The tree requires an inverted index structure in which each node stores a list of indexes. This list can be as long as the size of the training data, 
i.e., $\mathcal{O}(N)$. Overall, the memory requirement grows linearly in accordance to the training data.

In practice, SEQL almost never reaches the worst scenario. When using symbolic transformations, repeated words can be discarded (often referred to as numerosity reduction in related literature \citep{lin-sax:dmkd03}). 
The pruning technique in SEQL is effective in practice and drastically diminishes the number of subsequences to be evaluated. Finally, in the case of multiple representations parallelism is possible since the individual training of models in a representation are independent.

\begin{table}[htb]
\begin{center}
\caption{Theoretical complexity of state-of-the-art time series classifiers.}
\label{table:complexity}
\renewcommand{\arraystretch}{1.5}
\begin{tabular}{ll}
\hline 
Algorithm & Theoretical Complexity \\ 
\hline 
mtSAX-SEQL & $\mathcal{O}(NL^{\frac{3}{2}}\log{}L)$ \\ 
mtSFA-SEQL & $\mathcal{O}(NL^{\frac{3}{2}})$ \\ 
BOSS & $\mathcal{O}(N^2 L^2)$ \\ 
BOSS VS & $\mathcal{O}(NL^{\frac{3}{2}})$ \\ 
WEASEL & $\mathcal{O}(N L^2)$ \\ 
EE\_PROP & $\mathcal{O}(N^2 L^2)$ \\ 
COTE & $\mathcal{O}(N^2 L^4)$ \\ 
Fast Shapelet & $\mathcal{O}(N L^2)$ \\ 
Learning Shapelet & $\mathcal{O}(N L^2)$ \\ 
\hline 
\end{tabular} 
\renewcommand{\arraystretch}{1}
\end{center}
\end{table}


\section{Evaluation}
\label{sec:experiments}

We study two symbolic representations (i.e., SAX and SFA) and evaluate new variants of SEQL-based algorithms for the TSC task. 
To test our approaches, we experiment in total with 8 different combinations between the two symbolic representations and the SEQL variants (Table~\ref{table:exp_combi}).
The parameter settings for the experiments are informed by existing literature on symbolic representations for TSC \citep{lin-sax:dmkd07,schaefer:dmkd16,Schafer:2017:weasel,fvseql:7930038} and are set as shown in Table~\ref{table:params}. 
For multiple representations we use a minimum sliding window of size $l=16$ and an increment step of $\sqrt{L}$. We set $w=16, \alpha=4$ for SAX as by \cite{fvseql:7930038}, 
and $w=8, \alpha=4$ for SFA, as by \cite{schaefer:dmkd16}.
For SEQL we use default parameters as by \cite{ifrim-seql:kdd11}.


\begin{table}[h]

\begin{center}
\caption{Parameter settings for the experiments: $l$ is the window size, $w$ is the word size and $\alpha$ is the alphabet size.}
\label{table:params}
\renewcommand{\arraystretch}{1.5}
\begin{tabular}{llll}
\hline 
Input representation(s) & $l$ & $w$ & $\alpha$ \\ 
\hline 
Single SAX & fixed: $0.2*L$ & 16 & 4 \\ 
Single SFA & fixed: $0.2*L$ & 8 & 4 \\ 
\hline 
Multiple SAX & varied: minl=16, increment $\sqrt{L}$ & 16 & 4 \\ 
Multiple SFA & varied: minl=16, increment $\sqrt{L}$ & 8 & 4 \\ 
\hline 
\end{tabular} 
\renewcommand{\arraystretch}{1}
\end{center}
\end{table}




Our proposed algorithms were tested on all 85 datasets of the UCR Time Series Classification Archive~\citep{ucr:keogh15}. The archive has been incrementally extended by researchers working with time series and contains a vast collection of data from multiple domains and problem types. 
It is perhaps the most common used benchmark for recent studies on TSC. We performed all experiments with the default single split of training and test set given by the benchmark.
Our test system is a Linux PC with Intel Core i7-4790 Processor (Quad Core HT, 3.60GHz), 16GB 1600 MHz memory and 256 Gb SSD storage. All our code was developed in C\texttt{++} and can be found at \url{https://github.com/lnthach/Mr-SEQL}.


Following are short recaps of all state-of-the-art algorithms included in the next sections.
Footnotes indicate from where we obtained the results. 
In some cases, we were unable to obtain the full results on all UCR datasets from the original authors, hence we opted to use the reproduced results from more recent studies.
\begin{itemize}
\item \textbf{COTE} (Flat-COTE~\citep{bagnall2015time} and HIVE-COTE~\citep{hive-cote}) are large ensembles of different time series classifiers\footnote{\label{ueasource}\url{http://www.timeseriesclassification.com/results.php}}. 
\item \textbf{BOSS and WEASEL} \citep{schafer2015boss,Schafer:2017:weasel} are multi-resolution SFA-based time series classifiers\footnote{\label{sfasource}\url{https://www2.informatik.hu-berlin.de/~schaefpa/weasel/}}. 
\item \textbf{Deep Learners} ~\citep{IsmailFawaz2018deep} include FCN, ResNet, MLP, Encoder, Time-CNN, MCDCNN, MCNN, t-LeNet and TWIESN\footnote{\url{https://github.com/hfawaz/dl-4-tsc}}.

\item \textbf{LS} (Learning Shapelet)~\citep{grabocka-lts:kdd14} learns shapelets from time series by optimizing an objective function\footnotemark[\getrefnumber{sfasource}].
\item \textbf{EE\_PROP}~\citep{Lines2015:ee} is an ensemble of distance-based classifiers\footnotemark[\getrefnumber{ueasource}].
\item \textbf{TSBF}~\citep{baydogan:tsbf} uses Random Forest on a generated bag of features\footnotemark[\getrefnumber{ueasource}].
\item \textbf{DTW and DTW\_CV}~\citep{Lines2015:ee} are 1NN classifiers with DTW as distance measurement. DTW\_CV uses cross validation to set a constraint for the warping window\footnotemark[\getrefnumber{sfasource}].
\item \textbf{ST} (Shapelet Transform)~\citep{Lines:2012:STT:2339530.2339579} transforms the data to shapelet space to improve classification accuracy\footnotemark[\getrefnumber{ueasource}].
\item \textbf{FS}(Fast Shapelet)~\citep{fast-shapelet:sdm13} finds $k$-best shapelets in the dimension-reduced space of SAX\footnotemark[\getrefnumber{ueasource}].
\item \textbf{SAX-VSM}~\citep{senin-saxvsm:icdm13} generates a dictionary of SAX words for each class. It tunes SAX parameters with an optimization algorithm\footnotemark[\getrefnumber{ueasource}].
\end{itemize}

For comparison of multiple state-of-the-art classifiers, we follow the recommendations by~\cite{Demsar:2006,garcia-extension,JMLR:v17:benavoli16a}: we first detect the significant differences in rankings with the Friedmann test and follow with the Wilcoxon signed rank test with Holm's correction and used the Critical Difference diagram (CD) for visualization (e.g., Figure~\ref{fig:cd_seql}). 
This type of diagram is used very often in the literature on TSC for visual comparison of different methods across different datasets \citep{bagnall2016great,schafer2015boss}. 
The diagram shows the classifiers on a spectrum of average error ranking (the rank of the method with regard to classification error, averaged across 85 datasets), 
therefore classifiers to the left of the diagram (lower rank) perform better than classifiers to the right. 
The cliques (thick horizontal black lines) group methods that do not have a statistically significant difference in performance. All procedures were performed using the tool \textit{scmamp}~\citep{scmamp}.

\subsection{Comparison of Algorithms}

To evaluate our approach, we first look at how the variants of SEQL-based classifiers fare against each other. 
Next we examine the group of SAX-based methods and then SFA-based methods. 
Finally, we compare our most accurate representative against the state-of-the-art.

\subsubsection{SEQL-based Methods}

\begin{figure}[!ht]
\centering
\includegraphics[width=1\textwidth]{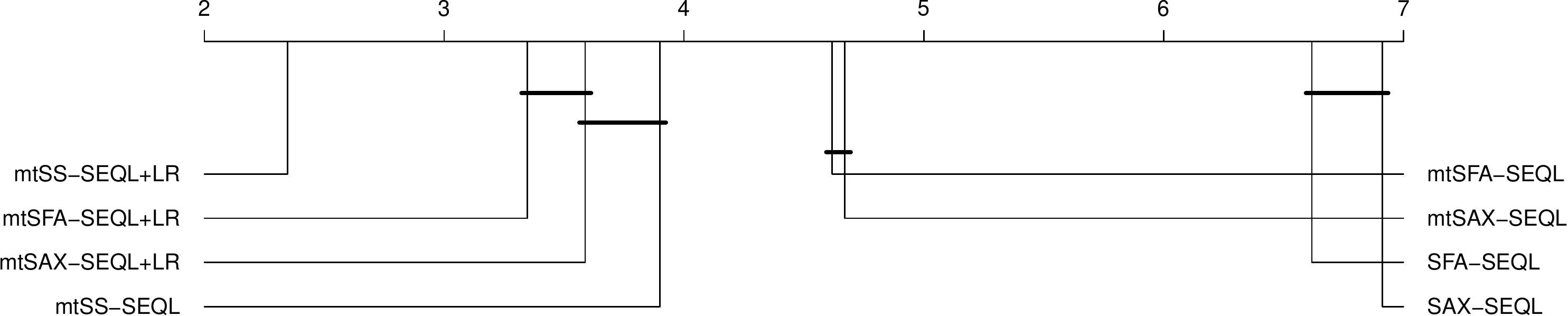}
\caption{Average error ranking of compared SEQL-based classifiers ordered (left-best) based on performance on the UCR Archive.
The best model mtSS-SEQL+LR is a linear model that combines features from SAX and SFA representations.}
\label{fig:cd_seql}
\end{figure}

Figure~\ref{fig:cd_seql} is a CD diagram that presents the ranking of  our proposed algorithms, based on the error attained on the UCR Archive.

As it can be seen, there is no critical difference between SAX-based classifiers (SAX-SEQL) and SFA-based classifiers (SFA-SEQL) even though SFA-based classifiers seem to perform better on average. 
The figure suggests that the substantial improvement in accuracy arrives from the combination of multiple representations, either by the ensemble of models or by combining features: 
the more symbolic representations are used, the more accurate the resulting classifier is. 
From right to left in the critical diagram (Figure~\ref{fig:cd_seql}), we start with a single representation (SAX/SFA), then add more representations of the same type (mtSAX/mtSFA), and finally combine representations of different types (mtSS). 
By combining representations from multiple-resolutions and multiple-domains to create features, we allow the classifier to select only those representations and features that represent the data well, 
and we do not have to decide in advance what are suitable symbolic parameters for the representations. We also note that the second algorithm we propose, which combines: (i) multiple symbolic representations, (ii)
feature selection with SEQL and (iii) a logistic regression classifier (mtSS-SEQL+LR) performs better than the ensemble algorithm (mtSS-SEQL). 
This is an interesting finding: creating a single rich feature space that combines symbolic representations and training a linear model delivers better accuracy than ensembling models trained with different representations.
The linear model also has the advantage that it is simple to interpret, as we discuss in Section \ref{sec:casestudy2}.

\subsubsection{SAX-based Methods}

\begin{figure}[!ht]
\centering
\includegraphics[width=0.9\textwidth]{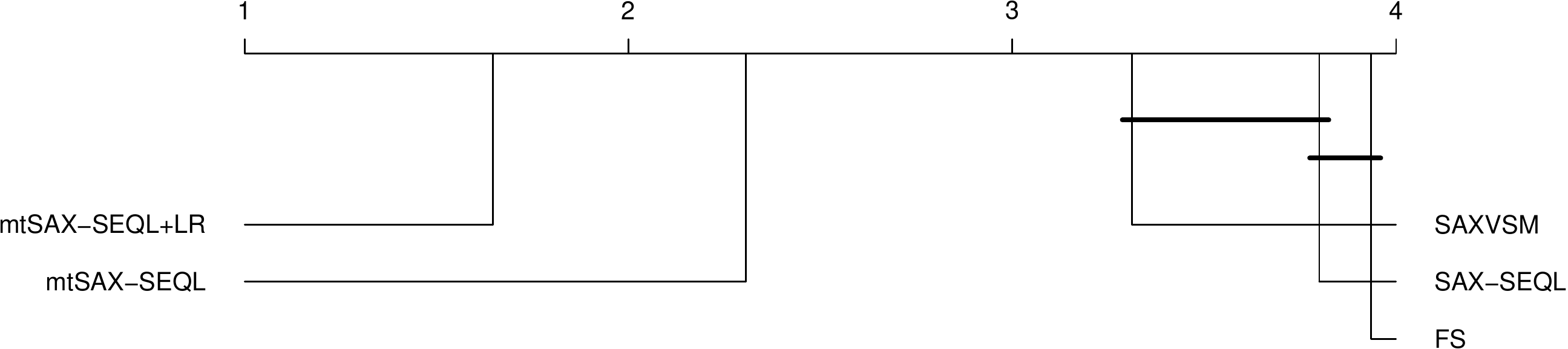}
\caption{Average error ranking of compared SAX-based classifiers ordered (left-best) based on performance on the UCR Archive.}
\label{fig:cd_sax}
\end{figure}

SAX-VSM and Fast Shapelet are perhaps the most well-known time series classifiers which utilize the SAX transformation. 
However, both have been already shown to be inferior in accuracy to more recent state-of-the-art methods. 
Our multi-resolution SAX-based variants also easily outperform both methods as can be seen in Figure~\ref{fig:cd_sax}.

From the diagram, we also notice that the difference between SAX-VSM and the single SAX-SEQL model is not critical, even though the former optimizes the parameters,  
while the latter uses a set of fixed parameter values. On the other hand, the low performance of Fast Shapelet may suggest the method fails 
to select the most discriminative features, despite the fact that it also employs multiple representations (multiple SAX resolutions).

\subsubsection{SFA-based Methods}

The SFA-based algorithm family includes WEASEL, BOSS and our SEQL-based classifiers (SFA-SEQL, mtSFA-SEQL and mtSFA-SEQL+LR). We excluded the mtSS variants since they incorporate not only SFA but also SAX representations.

\begin{figure}[!ht]
\centering
\includegraphics[width=0.9\textwidth]{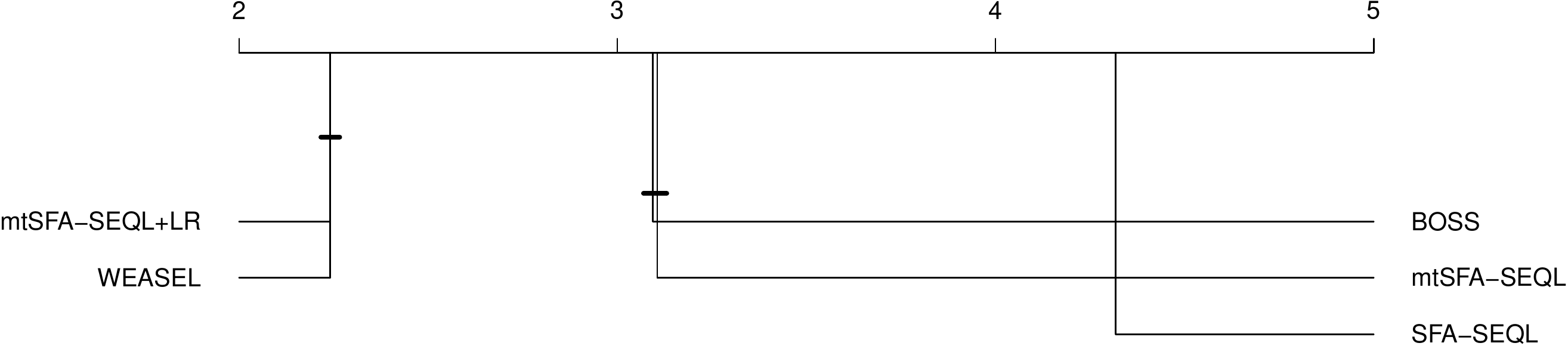}
\caption{Average error ranking of compared SFA-based classifiers ordered (left-best) based on performance on the UCR Archive.}
\label{fig:cd_sfa}
\end{figure}

Figure~\ref{fig:cd_sfa} shows the comparison between SFA-based classifiers. 
Interestingly, the performance of our algorithms matches those of BOSS (for mtSFA-SEQL) and WEASEL (for mtSFA-SEQL+LR) with almost no difference in terms of average ranking. It is worth noting that the output of the SFA transformation used in our experiments is more similar to the transformation used in BOSS, rather than in WEASEL. 
For WEASEL, the authors applied new transformation techniques including non-overlapping windows, bi-gram sequences and full ranges of sliding window sizes. 
The diagram highlights the potential advantage of our SEQL-based algorithms in exploring the all-subsequence SFA word space, an ability which the BOSS and WEASEL classifiers lack.

\subsubsection{Comparing to the State-of-The-Art}
\label{subsubsec:sota}

In this section, we investigate 22 time series classifiers including our representative mtSS-SEQL+LR. By picking DTW\_CV as the benchmark classifier (a common choice in literature~\citep{bagnall2016great,Lines2015:ee,Schafer:2017:weasel}), we divide the rest to three different groups based on the results of a Wilcoxon signed rank test at a cutoff of $p = 0.05$ (Table~\ref{table:dtw_diff}): significantly better, significantly worse or not significantly different to the benchmark.

\begin{table}[h]
\begin{center}
\caption{TSC algorithms grouped in comparison to DTW\_CV. The percentage in the brackets gives the difference in mean accuracy over the UCR Archive. 
Positive value means that this method is on average more accurate than the DTW\_CV benchmark and vice versa.}
\label{table:dtw_diff}
\renewcommand{\arraystretch}{1.5}
\begin{tabular}{lll}
\hline 
Significantly better & Not significantly different & Significantly worse \\ 
\hline 
HIVE-COTE (+8.7\%) & Encoder(-1.1\%) & DTW (-2.2\%) \\ 
mtSS-SEQL+LR (+7.9\%) & SAX-VSM (-1.8\%) & MLP (-3.4\%) \\  
Flat-COTE (+7.8\%) &  & Time.CNN (-4\%) \\  
WEASEL (+7.3\%) &  & FS (-5.3\%) \\ 
ResNet (+6.5\%) &  & MCDCNN (-7.8\%) \\ 
ST (+6.2\%) &  & TWIESN (-8\%) \\ 
BOSS (+5.4\%) &  & MCNN (-37.5\%) \\ 
FCN (+4.8\%) &  & t.LeNet (-39.8\%) \\ 
EE\_PROP (+3.3\%) &  &  \\ 
LearningShapelet (+2.7\%) &  &  \\ 
TSBF (+1.8\%) &  &  \\
\hline 
\end{tabular} 
\renewcommand{\arraystretch}{1}
\end{center}
\end{table}



\begin{figure}[!ht]
\centering
\includegraphics[width=1.05\textwidth]{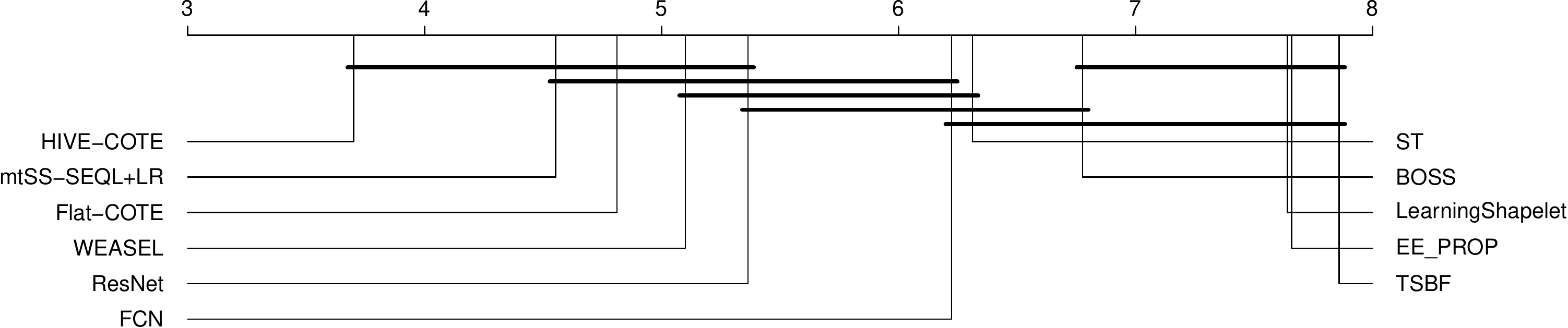}
\caption{Average error ranking of the most accurate classifiers ordered (left-best) based on performance on the UCR Archive.}
\label{fig:cd_best}
\end{figure}

Figure~\ref{fig:cd_best} shows the CD diagram for the classifiers which are found significantly better than the DTW\_CV benchmark. Ranking wise, HIVE-COTE takes the lead in average ranking followed by our method mtSS-SEQL+LR. Note that the COTE ensembles consist of time-based and frequency-based classifiers, hence they also rely on knowledge extracted from multiple domains. 
Nevertheless, the diagram also suggests that there is no significant difference between COTE classifiers and mtSS-SEQL+LR regarding their accuracy on the UCR benchmark.

Regarding time and space efficiency, FCN, ResNet and COTE are very demanding of computation resources and did not run on our machine.
We also found that WEASEL requires large amounts of memory to run, likely due to its enormous feature space. 
In our experiments on a regular Linux PC we got out-of-memory errors when we tried to run WEASEL even on 
moderately large datasets (e.g., a few hundred training time series).
Since these methods do not run on a regular PC (and many assume access to significant computation resources such as GPU clusters), 
 the accuracy of the state-of-the-art methods is reported from published results. 
 This also raises interesting research questions regarding the feasibility of 
 running many of the state-of-the-art TSC methods under strict resource-constrained requirements (e.g., running TSC on a mobile phone).


\begin{figure}[!ht]
\centering
\includegraphics[width=0.9\textwidth]{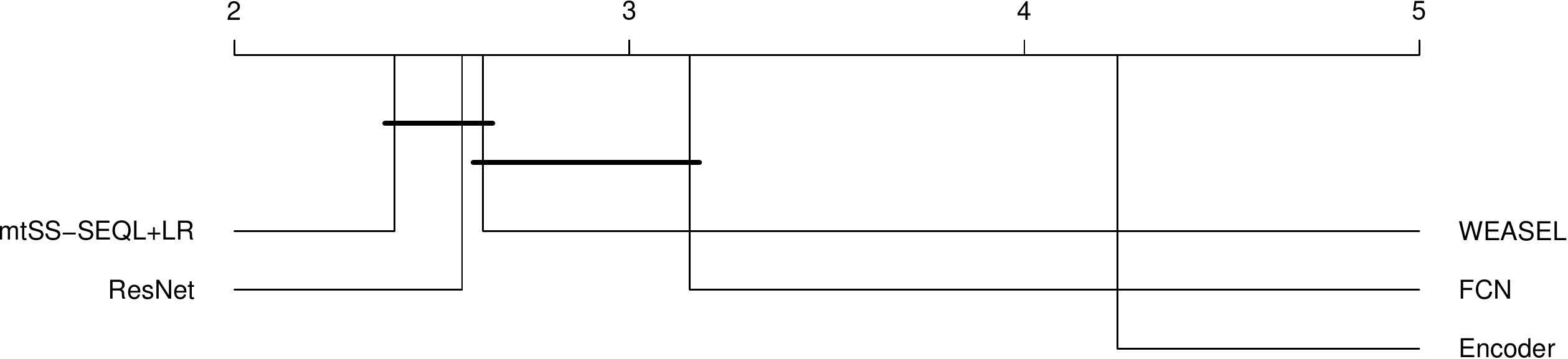}
\caption{Average error ranking of the most accurate linear classifiers and deep learning classifiers ordered (left-best) based on performance on the UCR Archive. 
Linear classifiers are much simpler but have comparable Accuracy to deep learners.}
\label{fig:cd_deep}
\end{figure}

In Figure~\ref{fig:cd_deep}, we examine the difference between the most accurate linear algorithms (mtSS-SEQL+LR and WEASEL) and deep learning algorithms. For deep learning algorithms, we select Encoder which is comparable to the benchmark (DTW\_CV) and FCN, ResNet~\citep{IsmailFawaz2018deep} which are significantly better than the benchmark. The diagram shows no significant difference between mtSS-SEQL+LR, ResNet, and WEASEL while both mtSS-SEQL+LR and ResNet are significantly better than FCN. 

\subsubsection{Comparing TSC Algorithms by Problem Type}

Following the discussion by \cite{bagnall2016great}, we also measured the performance of our proposed algorithms for each type of data in the UCR Archive. 
In summary, there are 7 types of data: Image Outline (29 datasets), Sensor Readings (16 datasets), Motion Capture (14 datasets), Spectrographs (7 datasets), Electric Devices (6 datasets), ECG measurements (7 datasets) and Simulated (6 datasets). 


\begin{table}[ht]
    \centering
    \caption{Average Error Rank by problem type for SEQL-based classifiers.}
   	\label{table:seql_rank_by_problem_type}
   	\renewcommand{\arraystretch}{1.4}
	\begin{tabular}{lccccccc}
	\hline 
	Methods & IMG & SPECTR & SENSOR & SIMUL & ECG & DEV & MOTION \\ 
	\hline 
	mtSS-SEQL+LR & \textbf{2.59} & \textbf{3.36} & \textbf{2.22} & \textbf{1.75} & \textbf{1.71} & 2.67 & \textbf{1.93} \\
	mtSFA-SEQL+LR & 3.24 & 3.93 & 3.16 & 3.67 & 2.57 & 4.83 & 3.11 \\
	mtSAX-SEQL+LR & 4.17 & 4.50 & 3.34 & 2.75 & 2.93 & \textbf{1.50} & 3.79 \\
	mtSS-SEQL & 4.16 & 4.57 & 3.94 & 4.25 & 4.00 & 3.17 & 3.11 \\
	mtSFA-SEQL & 4.17 & 5.43 & 4.66 & 4.67 & 4.57 & 6.25 & 4.39 \\
	mtSAX-SEQL & 4.53 & 3.50 & 5.28 & 4.58 & 5.21 & 3.33 & 5.18 \\
	SFA-SEQL & 6.38 & 4.93 & 6.56 & 7.33 & 7.36 & 7.67 & 6.89 \\
	SAX-SEQL & 6.76 & 5.79 & 6.84 & 7.00 & 7.64 & 6.58 & 7.61 \\
	\hline 
	\end{tabular} 
	\renewcommand{\arraystretch}{1}
\end{table}

Table \ref{table:seql_rank_by_problem_type} reports the average error rank of each SEQL-based method on each type of data. SFA-based classifiers appear to not be suitable for time series data from Electric Devices. In addition, we can also see the difference between SAX and SFA from the table, by comparing mtSAX-SEQL+LR and mtSFA-SEQL+LR (same classifier but with different input representation). It seems SFA has an advantage in Motion and Image categories while SAX has an advantage in Electric Devices category. 


\begin{figure}[!ht]
\centering
\includegraphics[width=\textwidth]{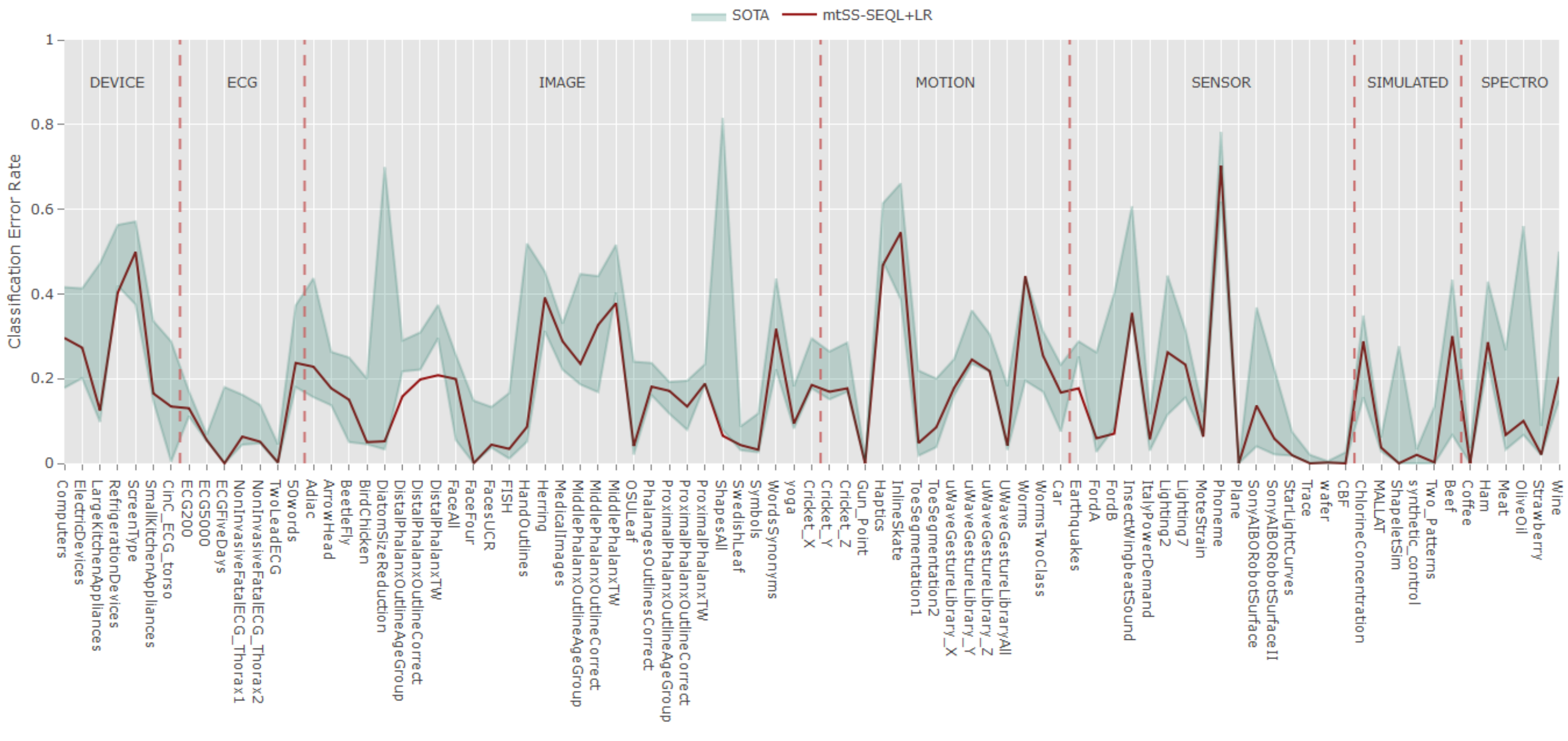}
\caption{Classification error rate of mtSS-SEQL+LR on 85 UCR Archive datasets. The light-colored area depicts the overall performance of the other ten state-of-the-art classifiers shown in the first column of Table~\ref{table:dtw_diff}. }
\label{fig:seql_vs_sota_error}
\end{figure}

Figure~\ref{fig:seql_vs_sota_error} provides an overview of mtSS-SEQL+LR (our best variant) performance in each group of problems compared to other state-of-the-art classifiers. To analyze the statistical differences, we performed the same test done in Section~\ref{subsubsec:sota} but separately for each type of data. Furthermore, we chose our classifier (mtSS-SEQL+LR) to be the benchmark this time. We found that, with the exception of Motion data, where HIVE-COTE is found to be significantly better, state-of-the-art classifiers are either significantly worse or not different to ours (Table~\ref{table:by_type}). The table also notes the difference in average error between mtSS-SEQL+LR and the corresponding method (negative difference means mtSS-SEQL+LR is more accurate). The test also suggests that it is harder to establish the difference between classifiers when the sample size is small, as in the case of problems with fewer datasets.

\begin{table}[h]
\begin{center}
\caption{TSC algorithms grouped in comparison to mtSS-SEQL+LR for each problem type. 
The percentage in the brackets gives the difference in mean accuracy over all datasets of the respective type. 
Positive value means that this method is on average more accurate than the benchmark (mtSS-SEQL+LR) and vice versa.}
\label{table:by_type}
\renewcommand{\arraystretch}{1.8}
\begin{tabular}{lll}
\hline 
Data type & Significantly better & Not significantly different\\ 
\hline 
 Image &  & HIVE-COTE (+0.3\%), WEASEL (-0.6\%), Flat-COTE (-1.1\%) \\ 
 (29 datasets) &  & ResNet (-4\%), FCN (-4.6\%) \\  
\hline 
 Sensor &  & Flat-COTE (+0.8\%), ResNet (+0.5\%), HIVE-COTE (-0.4\%), FCN (-0.9\%), \\ 
 (16 datasets) &   & LearningShapelet (-0.9\%), ST (-1.2\%), WEASEL (-2.3\%) \\ 
\hline
 Motion & HIVE-COTE (+1.5\%) & Flat-COTE (+1\%), WEASEL (+0.8\%), ST (-0.3\%),  \\
 (14 datasets) &  & BOSS (-1.6\%) \\
\hline
 ECG &  & Flat-COTE (+1.9\%), HIVE-COTE (+1.4\%), WEASEL (+1.3\%), \\
 (7 datasets) &  & ST (0\%), FCN (-0.2\%), ResNet (-0.9\%), Encoder (-1.4\%), \\
  &  &  BOSS (-1.9\%), EE\_PROP (-3.2\%), MLP (-4.5\%) \\
\hline 
 Spectro &  & HIVE-COTE (+2.3\%), WEASEL (+1.1\%), ST (+0.5\%), ResNet (+0.2\%), \\  
 (7 datasets) &  &  BOSS (-1\%), Flat-COTE (-1.3\%), SAXVSM (-1.9\%), \\ 
  &  & TSBF (-5.2\%), Time.CNN (-11.1\%) \\ 
\hline 
 Simulated &  & WEASEL (+0.5\%), HIVE-COTE (+0.3\%), Flat-COTE (-0.4\%), ResNet (-1.1\%), \\ 
 (6 datasets) &  & ST (-1.3\%), LearningShapelet (-1.9\%), TSBF (-3.2\%), EE\_PROP (-4.2\%), \\ 
  &  & FCN (-5\%), DTW (-7.2\%), MLP (-10.6\%), Encoder (-12.2\%) \\ 
\hline
 Device &  & HIVE-COTE (+2.6\%), ResNet (+2.3\%), FCN (+1.7\%), ST (-0.1\%), \\
 (6 datasets) &  & Flat-COTE (-1.2\%), BOSS (-3.8\%), SAXVSM (-4.9\%), WEASEL (-4.9\%), \\
  &  & TSBF (-10\%)  \\
\hline 
\end{tabular} 
\renewcommand{\arraystretch}{1}
\end{center}
\end{table}


We also note that all the datasets in UCR are sampled to fixed length, an artifact which hides the fact that most state-of-the-art methods do not work on variable-length time series. We elaborate more on this issue in Section \ref{sec:casestudy2}.

\subsection{Running Time of Our Algorithms}

Table \ref{table:seql_runtime} reports the average running time (in seconds) for each step in our experiments. As it was discussed previously, the independent training of each representation makes it possible to parallelise our algorithm. However, our current implementation is limited to sequential programming. Therefore, beside the total time for training (TotalLearn) and testing (TotalTest) we also report the longest time for training (MaxLearn) and testing (MaxTest) of a single representation per dataset, as the theoretical optimal running time for a parallel implementation. In summary, we show the runtime for the following steps in our algorithms:

\begin{itemize}
\item Transform: Average running time to transform raw data to symbolic representation. Note that we used the authors' implementation \citep{schaefer:dmkd16} for SFA transformation.
\item TotalLearn: Average training time in the case of multiple representations.
\item MaxLearn: Maximum average training time in the case of single representation.
\item TotalTest: Average testing time in the case of multiple representations.
\item MaxLearn: Maximum average testing time in the case of single representation.
\item LogReg: Average training and testing time with logistic regression.
\end{itemize}


\begin{table}[htb]
    \centering
    \caption{Average running time (seconds) of SEQL-based time series classifiers across all UCR datasets.}
   	\label{table:seql_runtime}
   	\renewcommand{\arraystretch}{1.4}
    \begin{tabular}{lcccccc}
    \hline 
    Methods & Transform & TotalLearn & MaxLearn & TotalTest & MaxTest & LogReg \\ 
    \hline 
    SAX-based & 83.219 & 131.530 & 3.704 & 67.748 & 2.554 & 5.829 \\
SFA-based & 4.210 & 66.127 & 2.233 & 18.036 & 0.729 & 5.941 \\
    \hline 
    \end{tabular} 
    \renewcommand{\arraystretch}{1}
\end{table}

We attempted to reproduce the experiments by other studies including SAX-VSM, BOSS, WEASEL, FCN and ResNet in order to have a fair comparison in terms of efficiency. However, except BOSS, none of these methods managed to complete the experiment on our machine (a regular PC). To the best of our knowledge the current implementation of BOSS is the most efficient implementation among the top accuracy classifiers. 
On the other hand, WEASEL demands a lot of memory which was simply not available on the machine we used. 
We suspect WEASEL's rich feature space and the absence of effective pruning techniques are the causes for this huge demand of computing resources.
In Section \ref{sec:casestudy2} we report the time and memory used by these algorithms to train and predict on a moderate-size human motion dataset.

\subsection{Comparing the Impact of Multiple Representations on Accuracy and Speed}
\label{mr-compare}
Previous experiments suggest that the strength of our models derives from multiple symbolic representations of time series. Adding representations seems to be the main factor for more accurate classification. However, it also raises computing cost as well as the risk of overfitting. We are also interested in the effect of combining representations from different domains. 

In this experiment, we adjust the number of representations and observe the impact on accuracy and running time. 
This can be done by expanding or shrinking the step when varying the window size $l$, e.g., instead of incrementing by $\sqrt{L}$, we can increment by $0.5 \sqrt{L}$ to increase the number of representations. 
The default configuration is similar to that of BOSS \citep{schafer2015boss}, i.e., varying different window sizes with a step of $\sqrt{L}$, as in the experiments for previous sections. The experiment was conducted with 3 variants of our SEQL-based classifier: mtSAX-SEQL+LR, mtSFA-SEQL+LR and mtSS-SEQL+LR.




\begin{figure}[!ht]
\centering
\includegraphics[scale=0.94]{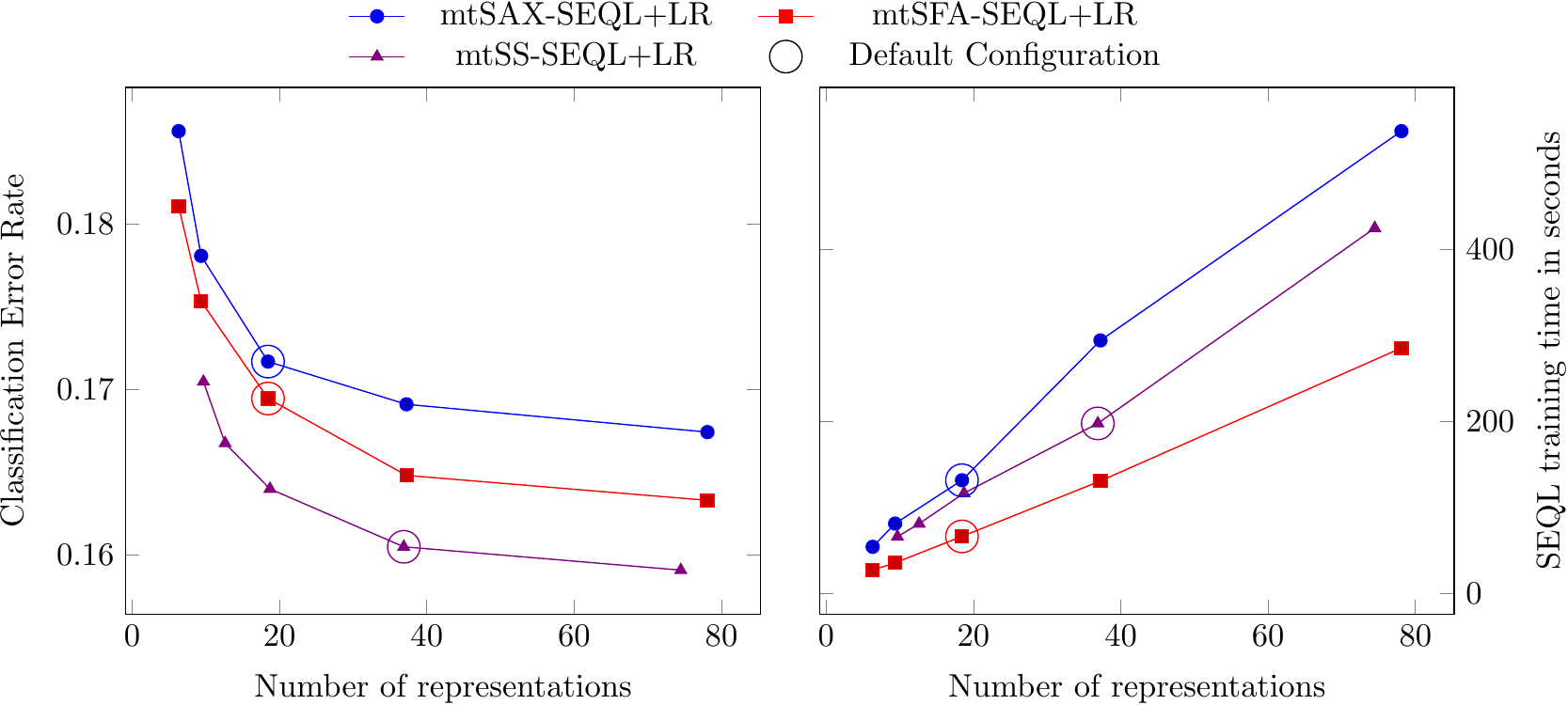}
\caption{ Impact of multiple symbolic representations on TSC: trade-off between accuracy and run time when increasing the number of input symbolic representations. 
All measurements are averaged across the UCR Time Series Archive.}
\label{fig:mtrep}
\end{figure}

Figure \ref{fig:mtrep} visualizes the results of the experiment. Indeed, feeding more representations to the classifier results in more accurate models, with extra cost however. In fact, our method can even achieve a lower error rate than the default configuration we used in the previous experiements. 
This experiment also suggests that multi-domain models (mtSS) are more accurate than single-domain models (mtSAX and mtSFA) given the same number of input representations (e.g., 20 representations for 
mtSS achieves lower error than 20 mt-SAX or 20 mt-SFA representations).


\section{Interpretability}
\label{sec:interpretability}
As described in Section~\ref{sec:method}, the output of our best SEQL-based algorithm is a linear model (a weighted list of selected features) which makes interpretation possible. 
In this section, we focus on classifier interpretation in the context of TSC, i.e., how can we identify the time series segments that are important for the classification decision. 
Since we visualize the data in the time domain, we only discuss SAX-SEQL-based models here. Regarding SFA representations, there have so far been no results that report whether SFA classifiers are interpretable. 
We visualize SAX sequences by mapping each SAX word back to its corresponding segment in the original time series. 
Technically, the same mapping can be done with SFA sequences, but currently we are not aware of good ways to visualize the impact of such (frequency domain) features, 
in particular since it is more intuitive to a human to visualise data in the time domain. 

\subsection{Feature Importance}

For our TSC algorithm mtSAX-SEQL+LR we can evaluate the importance of the features selected in the final model by studying the coefficients learned by logistic regression. 
Basically, the coefficient of a feature implies which class the feature represents (based on the sign) and how decisive the feature is in the classification decision (based on the absolute value). 
For multiple representations, a feature is defined not only by the sequence but also by its SAX parameters. 
Table~\ref{table:gun_point_top10} shows some of the features selected by the mtSAX-SEQL+LR algorithm on the GunPoint UCR time series dataset.



\begin{table}[htb]
    \centering
    \caption{Top 10 features selected by mtSAX-SEQL+LR from the GunPoint time series dataset.}
   	\label{table:gun_point_top10}
    \begin{tabular}{ccccl}
    \hline 
    \\[-0.7em]
    $l$ & $w$ & $a$ & Coefficients & Subsequences \\ 
    \\[-0.7em]
    \hline 
    \\[-0.7em]
    42 & 16 & 4 & 0.066 & cbaab \\
	53 & 16 & 4 & 0.062 & db \\
	53 & 16 & 4 & 0.062 & ddddb \\
	42 & 16 & 4 & 0.062 & da \\
	31 & 16 & 4 & 0.060 & bbbbbbbbbbcdddd \\
	53 & 16 & 4 & -0.054 & aaaaaabbbb \\
	20 & 16 & 4 & -0.054 & bbbbaaaaaa \\
	53 & 16 & 4 & -0.055 & bbbcddddd \\
	53 & 16 & 4 & -0.056 & bbbbbbbaaa \\
	53 & 16 & 4 & -0.061 & bbbbbbaaa \\
	\\[-0.7em]    
    \hline 
    \end{tabular} 
\end{table}

\subsection{Visualizing SAX Features}

\begin{figure}[h]
\begin{center}
\includegraphics[width=0.8\textwidth]{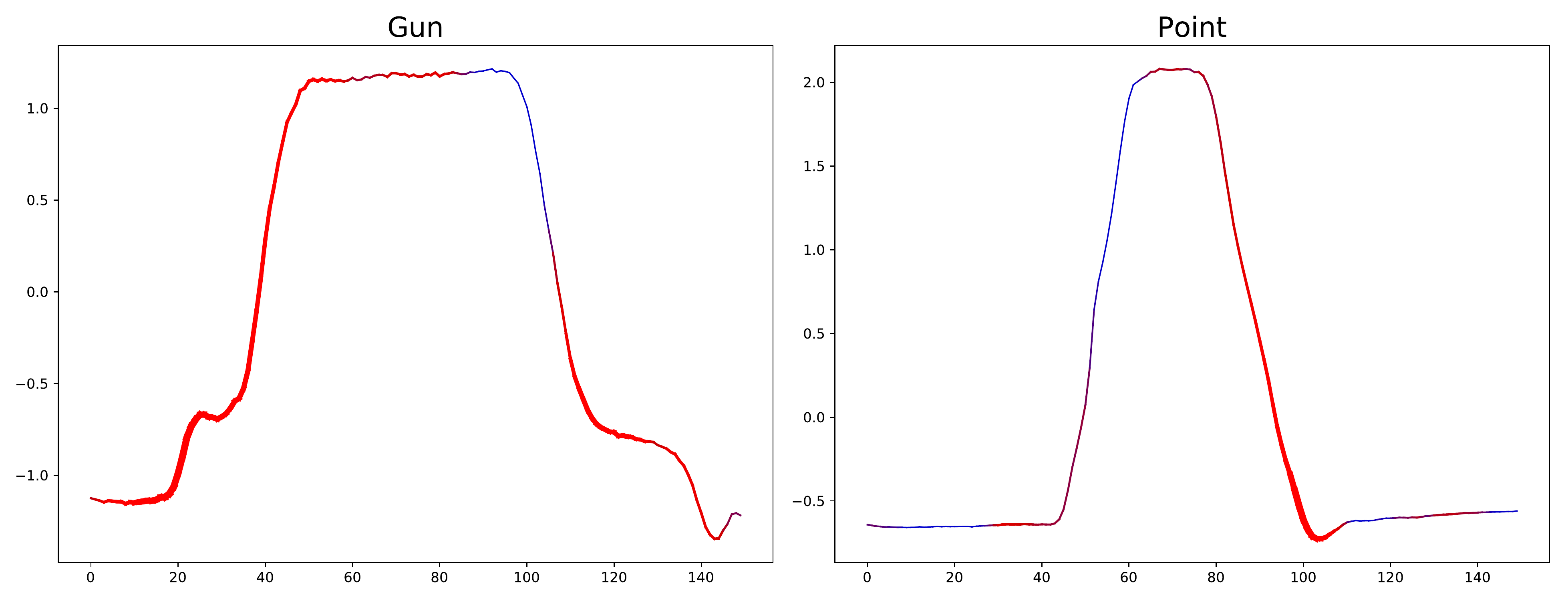}
\caption{Example from the GunPoint dataset. The discriminative regions for the respective class are thickened and highlighted with red color.}
\label{figure:gunpoint_example}
\end{center}
\end{figure}

Two examples of time series from the GunPoint dataset are shown in Figure~\ref{figure:gunpoint_example}. The time series records the motion of the hand when pointing (Point class) or drawing a gun (Gun class). The distinction between two time series can be observed at both the beginning, where a small bump describes gun-drawing action, and the end, where a little dip suggests the hand moved past the holster because there is no gun. The highlighted regions were discovered by our mtSAX-SEQL+LR classifier by mapping the matched features back to the raw time series (Algorithm~\ref{alg:mapsax}).

\begin{algorithm}[h]
\caption{Mapping symbolic features back to the original time series.}
\begin{algorithmic}[1]
\Function{findSegments}{feature, timeseries}
\State Initialize the meta time series $mtts = zeros(lengthof(timeseries))$ 
\State $sax = SAXtransform(timeseries,feature.l,feature.w,feature.\alpha)$  
\State Find all the locations of $feature.sequence$ in $sax$.
\ForAll{$loc$ in $locations$}
\State $mtts[loc] += feature.coef/locations.size()$
\EndFor
\State \Return $mtts$
\EndFunction
\end{algorithmic}
\label{alg:mapsax}
\end{algorithm}

\begin{figure}[h]
\begin{center}
\includegraphics[width=0.8\textwidth]{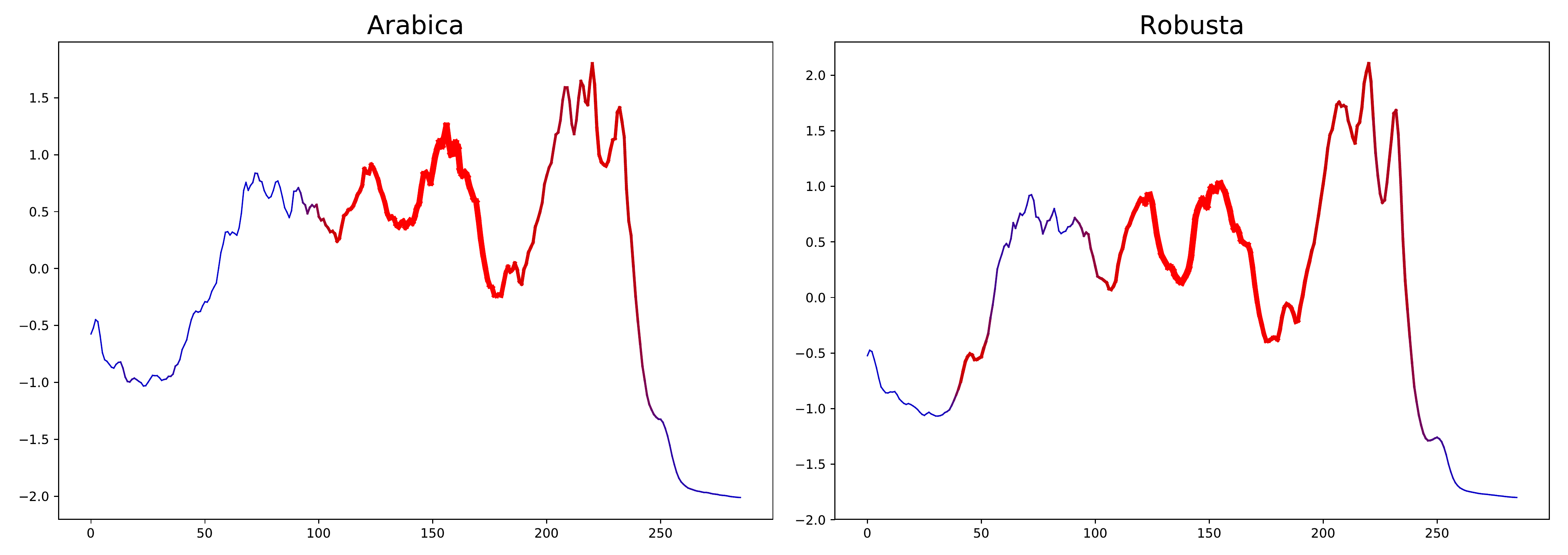}
\caption{Example from the Coffee dataset. The discriminative regions are thickened and highlighted with red color. }
\label{figure:coffee_example}
\end{center}
\end{figure}

Figure~\ref{figure:coffee_example} presents another two examples from the Coffee dataset: one from the Arabica class and one from the Robusta class. The highlighted regions correspond to the caffeine and chlorogenic acid components of the coffee blends~\citep{coffee}.

Our discoveries on the Coffee and GunPoint datasets have been well-documented by similar studies~\citep{ye-shapelets:kdd09,senin-saxvsm:icdm13,grabocka-lts:kdd14}. 
This suggests that our method is capable to identify the key discriminative regions of the time series. 
In the next section, we present a detailed case study in which we study the interpretability of our method on a real-world problem.


\section{Case Study: Human Motion Time Series Classification}
\label{sec:casestudy2}
To comprehensively investigate the interpretability of the proposed algorithms, we conducted a case-study on a dataset collected and well understood by the authors. 
The dataset describes the assessment of motion patterns during jumping performance testing in athletes. Below we describe the problem, the data collection process and 
the analysis of TSC algorithms.


\subsection{Problem Description}
The assessment of power output is an important component of athlete performance testing. 
Lower limb power is commonly assessed using the Countermovement Jumps (CMJ)  as it is a simple test 
and has been shown to be related to maximal and explosive strength performance \citep{Nuzzo2008RelationshipBC}. In order for a CMJ test to be considered acceptable, athletes must jump with adherence to acceptable technique. 
The athlete stands upright with weight evenly distributed over both feet. Hands are placed on the hips and stay there throughout the test. 
When ready, the athlete squats down until the knees are bent at 90 degrees, then immediately jumps vertically as high as possible, 
landing on both feet at the same time. Commonly observed deviations from acceptable technique, 
which indicate a failed test, include the athlete bending their legs in the air and stumbling upon landing \citep{markovic2004reliability}. 

Traditionally, CMJ tests are measured with large, cumbersome equipment such as force plates and infrared-based systems \citep{glatthorn2011validity}. 
More recently, it has been shown that accelerometer-based assessment of CMJ tests can improve equipment affordability and portability \citep{o2018wearable,picerno2011countermovement}. 
With the arrival of these new systems for measuring CMJ performance, it is important to ensure that they can automatically detect aberrant CMJ technique, 
which indicates a failed test, so that incorrectly completed tests are not saved as part of an athlete's performance profile. 
In this case-study we collected a dataset which captured acceptable CMJ tests and failed CMJ tests whereby the athlete bended their legs during flight or stumbled upon landing (3 classes of jumps: Normal, Bending, Stumble). 
The aims of the study were to compare the classification techniques presented in this paper to other methods commonly used in the literature, with regards to accuracy, time/space efficiency and interpretability. 

\subsection{Data Collection}
Ten participants (3 females, 7 males, age: 26.6 $\pm$2.2, weight: 80.1$\pm$7.4 kg , height: 1.8$\pm$0.1 m) were recruited for this case-study. 
The Human Research Ethics Committee at University College Dublin approved the study protocol  and written informed consent was obtained from all participants before testing. 
Participants did not have a current or recent musculoskeletal injury that would impair performance of CMJs. Participants were equipped with a Shimmer 3 (Shimmer, Dublin, Ireland) inertial measurement unit (IMU) 
on their dominant foot. The IMU was configured to stream wide range, tri-axial accelerometer data ($\pm$16 g) at 1024 Hz. 
Each participant completed 20 CMJs with acceptable form, 20 jumps with their legs bending during flight and 20 jumps with a stumble upon landing. 
The resulting 3-class dataset consists of 200 files of IMU data in the acceptable form class and, due to Bluetooth dropping twice during data collection, 
199 files of IMU data in each of the 'legs bending' and 'stumble on landing' classes. The length of the IMU time series signals in each file ranged from 1,231 to 6,710 samples.

\subsection{Data Analysis}
Acceleration magnitude was first computed from the accelerometer \textit{x, y} and \textit{z} signals  whereby: \[A_m=\sqrt{A^2_x + A^2_y+A^2_z}.\]
The signal used, as included in the data donated with this paper, was the acceleration magnitude 
inclusive of the inertial and gravitational accelerations acting on the sensor device.
This signal is appropriate for computing jump performance metrics, e.g., time in the air, jump height and classifying jump technique quality as described in this case-study. 
For the purposes of jump assessment, the removal of gravitational acceleration from the sensor's signal(s) is not important and has not been included in the methodology.
Classification was then completed using the methods introduced in this paper, state-of-the-art methods from the TSC literature and feature-based methods
 (Table \ref{table:jumpAccuracies}). 
The feature-based methods analysed were Support Vector Machines (SVM) and Random-Forests. 
The features used were mean, root-mean-square, standard deviation, kurtosis, median, skewness, range, variance, max value, minimum value, energy, 25th percentile, 75th percentile, level crossing rate, fractal dimension, index of minimum value, index of maximum value, number of signal peaks and length of signal epoch~\citep{OReilly2017ClassificationOD}. 
The Random-Forest method used 1000 trees. 
The SVM used a quadratic kernel function and a box constraint of 1.  
All classifiers were trained and evaluated using three settings. In the first setting, each full signal epoch was used (full signals data). 
In the second setting, the epoch was cropped whereby the active region of the jump was extracted (cropped signals data). 
This was completed by first computing the mean of the first 100 samples of the signal, 
the start point of the epoch was then set as the first value in the signal which was three times greater than the mean value. 
The end sample of the epoch was found by iterating backwards across the signal epoch and identifying the same threshold. 
In the third setting, all the cropped signal epochs were re-sampled to a fixed length of 500 samples (cropped and resampled to fixed length signals data).  
For this three-class (Normal, Bending, Stumble) classification problem the training set was the data from 70\% of the participants 
and the test set was data from 30\% of participants. No training and test data were from the same participants. 
The interpretability of all classification methods was considered by a domain expert in jump technique biomechanics and exercise classification with wearable sensors.

\subsection{Results}

The classification error rate, runtime, and memory usage for each technique are shown in Table \ref{table:jumpAccuracies}, \ref{table:jumpTime}, and \ref{table:jumpMem} respectively. 
For memory usage, we monitored the memory activity of the process with \textit{psrecord}\footnote{https://github.com/astrofrog/psrecord} and report only the highest peak. 
Note that we also tested HIVE-COTE for this case study, but the algorithm failed to complete the experiment in a reasonable amount of time on our system (we stopped it after 5h of training), 
hence we do not report results for this method.

When classifiers were trained and evaluated with the full signal epochs (variable-length time series), the mtSAX-SEQL+LR algorithm 
produced the lowest error rate\footnote{The error rates of mtSAX-SEQL+LR, mtSFA-SEQL+LR and mtSS-SEQL+LR were comparable, but we only discuss 
mtSAX-SEQL+LR here since it is also interpretable. Note that we used a default number of symbolic representations/resolutions for mtSAX-SEQL+LR (increasing the window length with a step $\sqrt{L}$). 
By increasing the number of SAX representations, we can further improve the accuracy of mtSAX-SEQL+LR, as also discussed in Section \ref{mr-compare}.}.
A number of methods could not be evaluated on the full signals data (F) as the varying length of the time series was not supported by the method or attempting to train the method 
caused out-of-memory errors. 
The error rate dropped for all methods when classifiers were trained and evaluated with the domain specific cropped signal epochs (C). 
In this scenario WEASEL has the lead, with the mtSAX-SEQL+LR, SVM and Random-Forest methods also performing well. 
In the final scenario, whereby the cropped signals were all re-sampled to a fixed length of 500 samples (CR), 
almost all methods achieved a very low error rate.

In terms of running time (Table \ref{table:jumpTime}), feature-based classifiers (SVM and Random Forest) have a clear advantage, with only BOSS VS as the only comparable method when the signals are cropped (in both C and CR columns). We argue that this advantage is mainly gained from the domain knowledge (to handcraft features), to which the other methods are oblivious. 
Deep learning methods (FCN and ResNet) took the longest time to finish the experiments. 
On the other hand, mtSAX-SEQL+LR shows its efficiency in memory usage (Table \ref{table:jumpMem}). In this category, the SFA-based family of classifiers (BOSS, BOSS VS, and WEASEL)
 reported very high memory activity. BOSS and WEASEL even failed to complete the experiments in the case of full signals (F). 
 Nevertheless, it is important to note that running time and memory activity strongly tie to the implementations of the algorithms, 
 which are usually written in different programming languages (C\texttt{++}, Java, Python, and Matlab in this case), hence might not fully reflect the efficiency of said algorithms.

\begin{table}[htb]
	\centering
	\caption{Classification error rate of each method compared on full signals (F), cropped signals (C) and cropped and re-sampled to fixed length signals (CR). 
	(-) The algorithm throws out-of-memory error in the experiment. (--) The implementation of the algorithm only supports fixed length time series. Best method in bold.}
	\label{table:jumpAccuracies}
	\begin{tabular}{lccc}
	\hline 
	\\[-0.8em]
	Methods & F & C & CR \\ 
	\\[-1em]
	\hline 
	\\[-0.7em]
	mtSAX-SEQL+LR & \textbf{0.123} & 0.056 & 0.028 \\
	WEASEL & (-) & \textbf{0.022} & \textbf{0.017} \\
	BOSS & (-) & 0.251 & 0.117 \\
	BOSS VS & 0.458 & 0.162 & 0.045 \\
	FCN & (--) & (--) & 0.045 \\
	ResNet & (--) & (--) & 0.067 \\
	SVM & 0.195 & 0.078 & 0.084 \\
	RandomForest & 0.251 & 0.061 & 0.039 \\
	1NN-DTW & (--) & (--) & 0.061 \\
	\\[-0.7em]
	\hline 
	\end{tabular} 
\end{table}

\begin{table}[htb]
	\centering
	\caption{Runtime in seconds of each method compared on full signals (F), cropped signals (C) and cropped and re-sampled to fixed length signals (CR). 
	(-) The algorithm throws out-of-memory error in the experiment. (--) The implementation of the algorithm only supports fixed length time series. Best method in bold.}
	\label{table:jumpTime}
	\begin{tabular}{lccc}
	\hline 
	\\[-0.8em]
	Methods & F & C & CR \\ 
	\\[-1em]
	\hline 
	\\[-0.7em]
	mtSAX-SEQL+LR & 1717.288 & 139.045 & 102.473 \\
	WEASEL & (-) & 688.523 & 307.247 \\
	BOSS & (-) & 724.522 & 406.285 \\
	BOSS VS & 127.658 & 17.161 & 13.133 \\
	FCN & (--) & (--) & 8900.759 \\
	ResNet & (--) & (--) & 14259.69 \\
	SVM & \textbf{13.1} & \textbf{11.05} & \textbf{10.51} \\
	RandomForest & 18.89 & 16.33 & 15.76 \\
	1NN-DTW & (--) & (--) & 1167.986 \\
	\\[-0.7em]
	\hline 
	\end{tabular} 
\end{table}

\begin{table}[htb]
	\centering
	\caption{Memory usage in MBs of each method compared on full signals (F), cropped signals (C) and cropped and re-sampled to fixed length signals (CR). 
	(-) The algorithm throws out-of-memory error in the experiment. (--) The implementation of the algorithm only supports fixed length time series. Best method in bold.}
	\label{table:jumpMem}
	\begin{tabular}{lccc}
	\hline 
	\\[-0.8em]
	Methods & F & C & CR \\ 
	\\[-1em]
	\hline 
	\\[-0.7em]
	mtSAX-SEQL+LR & 537.762 & \textbf{127.02} & \textbf{117.559} \\
	WEASEL & (-) & 2860.457 & 2398.332 \\
	BOSS & (-) & 2688.645 & 1973.254 \\
	BOSS VS & 3498.031 & 1552.613 & 1494.125 \\
	FCN & (--) & (--) & 801.609 \\
	ResNet & (--) & (--) & 892.578 \\
	SVM & \textbf{497.609} & 495.707 & 502.438 \\
	RandomForest & 528.016 & 519.609 & 520.039 \\
	1NN-DTW & (--) & (--) & 132.562 \\
	\\[-0.7em]
	\hline 
	\end{tabular} 
\end{table}

\begin{figure}[h]
	\begin{center}
		\includegraphics[width=\textwidth]{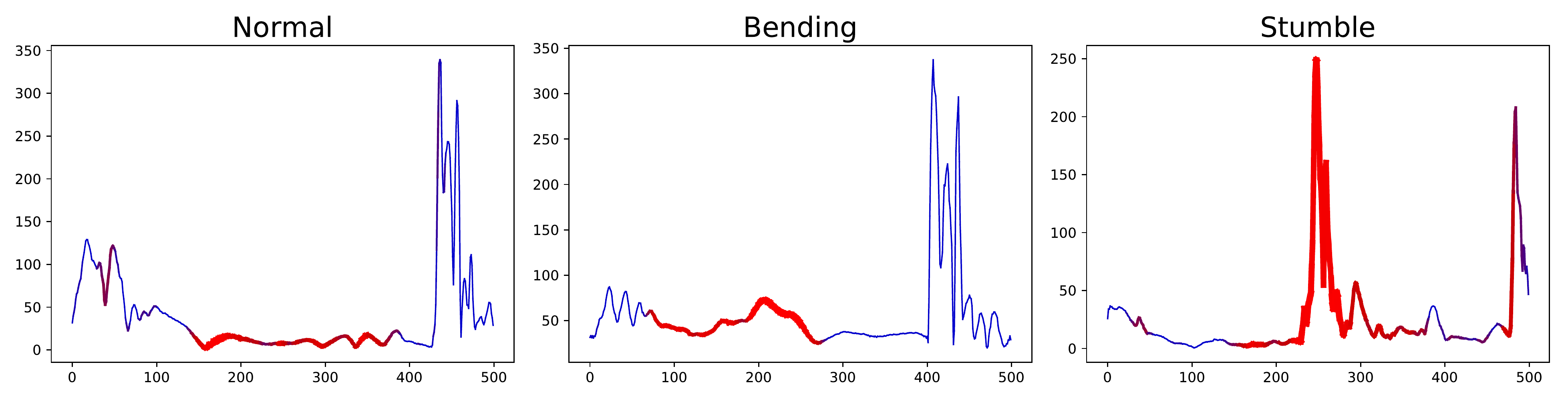}
		\caption{Three different motions made by a person in a CMJ test. Signals were cropped and re-sampled to fixed length. 
		The red-highlighted segments are features selected by the linear classifier mtSS-SEQL+LR and suggest the discriminative segments of the motion for the respective class.}
		\label{figure:jump_P1}
	\end{center}
\end{figure}

For each classifier evaluated, the interpretability of the method was considered by the domain expert. 
For the feature-based methods (SVM and RandomForest) and the other methods performing well (BOSS and WEASEL) the reasons for successful classification and for misclassification were largely unknown. 
The domain expert cited a key advantage of the mtSAX-SEQL+LR method to be  the ability to visualize the sub-sections of signals which are pertinent to differentiating the classes. 
For instance, this approach allows to clearly highlight the segment of the signal which represents a stumble upon landing (Figure~\ref{figure:jump_P1}). 
It also identified and highlighted the exact portion in the signal whereby participants were bending their legs in the air (Figure~\ref{figure:jump_P1}). 
This was considered particularly useful when the visualizations were placed side by side to a normal jump. 
This allowed the domain expert to quickly understand the differences in acceleration profiles between jumps completed with acceptable technique 
and with known deviations from correct technique. Additionally, the  mtSAX-SEQL+LR method was deemed useful for identifying less intuitive differences between classes. 
For instance the highlighted portion on the first landing spike in the stumble class (Figure~\ref{figure:jump_P1}) suggests that when an athlete lands correctly versus when they stumble, 
there is a different acceleration profile following initial contact. This was deemed new insight for the domain expert.  

\subsection{Discussion}

Whilst many of the methods evaluated in Table \ref{table:jumpAccuracies} produced favourable accuracy results following 
the pre-processing steps of cropping the signal to the region representing the jumping motion and then re-sampling the time series to a fixed length, 
it is interesting to see that the mtSAX-SEQL+LR method achieved acceptable accuracy when the classifier was trained and tested on the raw data. 
This suggests that this method has a better inherent ability to factor out irrelevant information in raw signals, 
than the feature-based methods, the BOSS/WEASEL or the deep learning FCN/ResNet methods. It is also of interest to note that the mtSAX-SEQL+LR method  
was able to perform training/testing on the varying lengths of signal epochs without the necessity of altering the data by padding or re-sampling the time series to a pre-defined fixed length. 

With regards to the interpretability of the presented methods, the domain expert stated that the visualizations from the mtSAX-SEQL+LR method 
contributed to an understanding of both how the classifiers actually worked and in understanding the differences in acceleration profiles between the various classes. They saw particular potential for the presented methods to educate athletes, using wearable sensor data, on the portions of their movements  which need correcting in order to perfect a skillful, complex movement, e.g., squatting exercise or tennis serve. 
All the data from this case-study was donated to the UCR and UEA Time Series Classification Repository and released together with the source code for this paper to allow reproducibility and to enable the development of accurate and interpretable methods on raw, real-world, time series problems and datasets.

\section{Conclusion}
\label{sec:conclusion}

The goal of this study is to explore the impact of combining multiple resolutions and multiple domains of time series symbolic representations with efficient sequence classifiers, 
while posing both accuracy and classifier interpretability as desirable properties.
In the TSC area it has been commonly perceived that DTW-based methods are hard to beat and that ensemble methods (e.g., COTE), and lately deep learning methods (e.g., FCN), 
offer the best accuracy. 
However, the series of SFA-papers \citep{Schafer:2012:SSF:2247596.2247656,schafer2015boss,schaefer:dmkd16,Schafer:2017:weasel} has shown otherwise.
In particular the WEASEL method is an accurate algorithm that uses a linear classifier in a large-dimensional SFA-words space. 
As discussed in our experiments, WEASEL faces memory challenges due to the large feature space it uses, and is not interpretable. 
Our concerns about interpretability and efficiency mean that even the most accurate classifiers can still be less desirable. 
With this paper, we have shown that linear classifiers are strong competitors with regard to accuracy, and that symbolic representations are a powerful tool for time series analysis. 

In summary, we have studied two notable symbolic representations of time series (SAX and SFA) and proposed a time series classification framework that can utilize both representations at different resolutions. 
Due to its pruning ability, our core classifier (SEQL) can navigate the vast symbolic-words space efficiently. We showed that symbolic approximation at multiple resolutions is an effective approach, 
instead of trying to find an optimal symbolic representation. Furthermore, our classifiers work with different symbolic representations, thus effectively learn from a multiple domain feature space 
without the need of incorporating various learning algorithms. 
We think that this characteristic has great potential as our classifier can theoretically accommodate other symbolic representations in the future. 
In practice, this flexibility means representations and resolutions can be chosen according to the problem and application domain.

We proposed 8 different SEQL-based algorithms using the SAX and SFA symbolic representations.
To showcase our contributions, we tested our proposals with the full UCR Time Series Archive and demonstrated that they are strongly competitive against the state-of-the-art,
 including against complex algorithms such as large ensembles (COTE) or deep learning methods (FCN). 
 While ensemble and deep learners are well-known for their accuracy, they are also notorious for their high demand of computing resources. 
 On the other hand, our SEQL-based methods are more efficient due to the effective combination of symbolic representations and sequence learning algorithm.
The time and space complexity of our algorithms also enables them to scale well for large datasets. 
The outcome of our best SEQL-based TSC algorithm is a linear model, which enables us to interpret the classification decision, a property which is desirable for time series analysis. 

Various ways of interpreting the resulting models were also discussed in this paper. In particular, we discussed interpreting the classification decision on well known UCR problems, 
as well as presented a case-study on athlete performance testing, where we can relate the algorithmic decisions to real-world domain knowledge. 
We have also reported and discussed the accuracy, time and space efficiency for all the methods evaluated during the case-study. We pointed out that existing TSC approaches 
require pre-processing of the raw signal and need extensive computation resources to achieve high accuracy. This opens interesting research directions in resource-constraint 
TSC where the classifiers have access to limited computation resources (e.g., limited memory) for training and prediction. 
Such applications are already common in the deployment of TSC on mobile phones,
and we intend to extend our methods to further improve their efficiency for this constrained setting.
For the future, we also see a lot of potential on extending these methods for multivariate time series classification problems, as well as continuing to work 
on human-friendly ways of explaining the classification decisions.


\section*{Acknowledgment}

We would like to thank the anonymous reviewers for their detailed and constructive feedback.
We would also like to gratefully acknowledge the work by researchers at University of California Riverside, USA (especially Eamonn Keogh and his team) 
and researchers at University of East Anglia, UK (especially Tony Bagnall and his team) 
and their effort in collecting, updating and making available the UCR and UEA time series classification benchmarks.
We want to thank all researchers in time series classification who have made their data, code and results open source and have helped the reproducibility of research methods in this area.
We acknowledge financial support for this work by Science Foundation Ireland (SFI) under grant number 12/RC/2289 (Insight Centre for Data Analytics).

\bibliographystyle{spbasic}

\end{document}